\documentclass[11pt]{article}

\usepackage[preprint]{acl}

\usepackage{times}
\usepackage{latexsym}

\usepackage{amsmath,amsfonts,amssymb}

\usepackage{array}
\usepackage{booktabs}
\usepackage{tabularx}
\usepackage{longtable}
\usepackage{makecell}
\usepackage{ragged2e}
\usepackage[table]{xcolor}
\usepackage{multirow}

\usepackage{enumitem}

\usepackage{graphicx}
\usepackage{float}
\usepackage{wrapfig}
\usepackage{newfloat}

\usepackage{algorithm}
\usepackage{algorithmic}

\usepackage{listings}

\usepackage{caption}
\usepackage{subcaption}
\usepackage{url}
\usepackage{comment}

\usepackage{hyperref}

\definecolor{figuregray}{gray}{0.9}
\usepackage[T1]{fontenc}
\usepackage[utf8]{inputenc}

\usepackage{microtype}

\usepackage{inconsolata}

\usepackage{graphicx}
\usepackage[table]{xcolor}

\title{\textsc{Condesion-Bench}: Conditional Decision-Making of Large Language Models in Compositional Action Space}

\author{Yeonjun Hwang~~~
 Sungyong Park~~~
 Minju Kim~~~
 Dongha Lee~~~
 Jinyoung Yeo\thanks{~~Corresponding author} \\\\
  Department of Artificial Intelligence, Yonsei University\\
 \texttt{\{hbhhyj, jinyeo\}@yonsei.ac.kr}\\
 }

\begin{document}
\maketitle
\definecolor{green}{RGB}{36, 214, 36}
\definecolor{red}{RGB}{235, 30, 30}
\definecolor{lightredshade}{HTML}{dea9a9}
\definecolor{lightgreenshade}{HTML}{bce3bd}
\definecolor{lightblueshade}{HTML}{cacbe8}
\definecolor{MyDarkBlue}{rgb}{0,0.08,1}
\definecolor{MyDarkGreen}{rgb}{0.02,0.6,0.02}
\definecolor{MyDarkRed}{rgb}{0.8,0.02,0.02}
\definecolor{MyDarkOrange}{rgb}{0.40,0.2,0.02}
\definecolor{MyPurple}{RGB}{111,0,255}
\definecolor{MyRed}{rgb}{1.0,0.0,0.0}
\definecolor{MyGold}{rgb}{0.75,0.6,0.12}
\definecolor{MyDarkgray}{rgb}{0.66, 0.66, 0.66}

\definecolor{MyYellow}{rgb}{254, 246, 170}
\definecolor{MyBlue}{rgb}{170, 217, 251}
\definecolor{LuneBlue}{rgb}{0.11, 0.11, 0.43}

\newcommand{\greencheck}{\textcolor{green}{\ding{51}}}
\newcommand{\redcross}{\textcolor{red}{\ding{55}}}

\newcommand{\jy}[1]{\textcolor{MyDarkBlue}{[Jinyoung: #1]}}
\newcommand{\sy}[1]{\textcolor{MyDarkBlue}{[Shunyu: #1]}}
\newcommand{\hj}[1]{\textcolor{MyPurple}{[HJ: #1]}}
\newcommand{\sj}[1]{\textcolor{LuneBlue}{[SJ: #1]}}

\newcommand{\mcal}[1]{{\cal{#1}}}
\newcommand{\calA}{\mbox{${\cal A}$}}
\newcommand{\calB}{\mbox{${\cal B}$}}
\newcommand{\calC}{\mbox{${\cal C}$}}
\newcommand{\calD}{\mbox{${\cal D}$}}
\newcommand{\calE}{\mbox{${\cal E}$}}
\newcommand{\calF}{\mbox{${\cal F}$}}
\newcommand{\calG}{\mbox{${\cal G}$}}
\newcommand{\calH}{\mbox{${\cal H}$}}
\newcommand{\calI}{\mbox{${\cal I}$}}
\newcommand{\calJ}{\mbox{${\cal J}$}}
\newcommand{\calK}{\mbox{${\cal K}$}}
\newcommand{\calL}{\mbox{${\cal L}$}}
\newcommand{\calM}{\mbox{${\cal M}$}}
\newcommand{\calN}{\mbox{${\cal N}$}}
\newcommand{\calO}{\mbox{${\cal O}$}}
\newcommand{\calP}{\mbox{${\cal P}$}}
\newcommand{\calQ}{\mbox{${\cal Q}$}}
\newcommand{\calR}{\mbox{${\cal R}$}}
\newcommand{\calS}{\mbox{${\cal S}$}}
\newcommand{\calT}{\mbox{${\cal T}$}}
\newcommand{\calU}{\mbox{${\cal U}$}}
\newcommand{\calV}{\mbox{${\cal V}$}}
\newcommand{\calW}{\mbox{${\cal W}$}}
\newcommand{\calX}{\mbox{${\cal X}$}}
\newcommand{\calY}{\mbox{${\cal Y}$}}
\newcommand{\calZ}{\mbox{${\cal Z}$}}

% \definecolor{darkgreen}{RGB}{0,100,0}

\newcommand{\todocccc}[2]{{\textcolor{#1}{[[#2]]}}}
\newcommand{\todogreen}[1]{\todocccc{green}{[[#1]]}}
\newcommand{\haejuu}[1]{\todogreen{haejuu: #1}}
\newcommand{\yeo}[1]{\textcolor{purple}{#1}}
\newcommand{\dragon}[1]{\textcolor{brown}{#1}}
\newcommand{\bwoo}[1]{\textcolor{olive}{#1}}
\newcommand{\kyle}[1]{\textcolor{blue}{#1}} 
\newcommand{\cheris}[1]{\textcolor{teal}{#1}}
\newcommand{\chatgpt}[1]{\textcolor{gray}{#1}}
\newcommand{\minus}[1]{\textcolor{red}{#1}}
\newcommand{\plus}[1]{\textcolor{ForestGreen}{#1}}
\newcommand{\person}{{$\mathbb{P}$}}
\newcommand{\thought}{$\mathbb{Z}_{\mathbb{P}}$}
\newcommand{\ty}[1]{\textcolor{darkgreen}{[TY: #1]}}

\newcommand{\se}{{\it SE}}%
\newcommand{\eg}{{\it e.g.},~}%
\newcommand{\ie}{{\it i.e.},~}%
\newcommand{\etal}{{\it et al.}}%
\newcommand{\etc}{{\it etc}}%

\newcommand{\worldmodel}{$\mathcal{W}_{\theta}$}
\newcommand{\ours}{\textsc{Think-and-Execute}\xspace}
\newcommand{\coffeegym}{\textsc{Coffee-Gym}\xspace}
\newcommand{\cf}{\textsc{Coffee}\xspace}
\newcommand{\coffeeeval}{\textsc{CoffeeEval}\xspace}
\newcommand{\editeval}{\textsc{CoffeeEval}\xspace}
\newcommand{\editevalbf}{\textbf{\textsc{CoffeeEval}}\xspace}
\newcommand{\coffeewemoji}{\coffee\xspace\cf}
\newcommand{\coffeewemojibf}{\coffee\xspace\textbf{\cf}}
\newcommand{\cfwemoji}{\coffee\xspace\cf}
\newcommand{\cp}{\textsc{CoffeePots}\xspace}
\newcommand{\coffeepots}{\textsc{CoffeePots}}
\newcommand{\argmin}{\operatornamewithlimits{argmin}}
\newcommand{\argmax}{\operatornamewithlimits{argmax}}

\newcommand{\blueText}[1]{\textcolor{blue}{#1}}
\newcommand{\greenText}[1]{\textcolor{darkgreen}{#1}}
% \definecolor{pythonblue}{rgb}{0.16,0.12,0.93}
% \definecolor{cppgreen}{rgb}{0.16,0.42,0.16}
% \definecolor{promptinsert}{HTML}{bfefff}
% \definecolor{compcolor}{HTML}{90EE90}
% \definecolor{codehlcolor}{HTML}{ffec8b}
% \definecolor{codehlcolor2}{HTML}{ffbbff}
% \definecolor{bgcolor}{rgb}{0.95,0.95,0.92}

\lstdefinestyle{python}{
    language=Python,
    basicstyle=\fontsize{8}{10}\ttfamily,
    keywordstyle=\color{blue},
    commentstyle=\color{gray},
    stringstyle=\color{black},
    showstringspaces=false,
    breaklines=true,
    breakindent=0pt,
    breakatwhitespace=false,
    escapeinside={(*@}{@*)}
}

\lstdefinestyle{cpp}{
    language=C++,
    basicstyle=\fontsize{8}{10}\ttfamily,
    keywordstyle=\color{blue},
    commentstyle=\color{gray},
    stringstyle=\color{green},
    showstringspaces=false,
    breaklines=true,
    breakindent=0pt,
    breakatwhitespace=false,
    escapeinside={(*@}{@*)}
}

% Small styles for examples in main text
\lstdefinestyle{plain}{
    basicstyle=\fontsize{8}{10}\ttfamily,
    keywordstyle=\color{blue},
    commentstyle=\color{gray},
    stringstyle=\color{green},
    showstringspaces=false,
    breaklines=true,
    breakatwhitespace=false,
    breakindent=0pt,
    escapeinside={(*@}{@*)}
}

\lstdefinestyle{python2}{
    language=Python,
    basicstyle=\fontsize{8}{10}\ttfamily,
    keywordstyle=\color{blue},
    commentstyle=\color{gray},
    stringstyle=\color{green},
    showstringspaces=false,
    breakatwhitespace=false,
    breaklines=true,
    breakindent=0pt,
    escapeinside={(*@}{@*)}
}

\lstdefinestyle{cpp2}{
    language=C++,
    basicstyle=\fontsize{8}{10}\ttfamily,
    keywordstyle=\color{blue},
    commentstyle=\color{gray},
    stringstyle=\color{green},
    showstringspaces=false,
    breaklines=true,
    breakindent=0pt,
    breakatwhitespace=false,
    escapeinside={(*@}{@*)}
}

\lstdefinestyle{sql}{
    language=SQL,
    basicstyle=\fontsize{8}{10}\ttfamily,
    keywordstyle=\color{blue},
    commentstyle=\color{green},
    stringstyle=\color{black},
    showstringspaces=false,
    breakatwhitespace=false,
    breaklines=true,
    breakindent=0pt,
    escapeinside={(*@}{@*)}
}

% Styles for prompts
% Prompt is for code, text is for regular text
\lstdefinestyle{prompt}{
    language=Python,
    basicstyle=\fontsize{8}{10}\ttfamily,
    keywordstyle=\color{blue},
    commentstyle=\color{gray},
    stringstyle=\color{cppgreen},
    showstringspaces=false,
    breaklines=true,
    backgroundcolor=\color{bgcolor},
    keepspaces=true, 
    breakindent=0pt,
    % linecolor=\color{lightgray},
    breakatwhitespace=false,
    showspaces=false,   
    escapeinside={(*@}{@*)}
}
\lstdefinestyle{text}{
    basicstyle=\fontsize{8}{10}\ttfamily,
    showstringspaces=false,
    breaklines=true,
    backgroundcolor=\color{bgcolor},
    breakatwhitespace=false,
    breakindent=0pt,
    keepspaces=true,
    showspaces=false,   
    escapeinside={(*@}{@*)}
}

\newcommand{\inserthl}[1]{\sethlcolor{promptinsert}\hl{#1}}
\newcommand{\comphl}[1]{\sethlcolor{compcolor}\hl{#1}}
\newcommand{\codehl}[1]{\sethlcolor{codehlcolor}\hl{#1}}
\newcommand{\codehlerr}[1]{\sethlcolor{codehlcolor2}\hl{#1}}

% \definecolor{lightblue}{RGB}{224,236,247}
% \definecolor{deepblue}{RGB}{9,46,107}
\begin{abstract}
Large language models have been widely explored as decision-support tools in high-stakes domains due to their contextual understanding and reasoning capabilities. However, existing decision-making benchmarks rely on two simplifying assumptions: actions are selected from a finite set of pre-defined candidates, and explicit conditions restricting action feasibility are not incorporated into the decision-making process. These assumptions fail to capture the compositional structure of real-world actions and the explicit conditions that constrain their validity. To address these limitations, we introduce \textsc{Condesion-Bench}, a benchmark designed to evaluate conditional decision-making in compositional action space. In \textsc{Condesion-Bench}, actions are defined as allocations to decision variables and are restricted by explicit conditions at the variable, contextual, and allocation levels. By employing oracle-based evaluation of both decision quality and condition adherence, we provide a more rigorous assessment of LLMs as decision-support tools.
% Large language models have been widely explored as decision-support tools in high-stakes domains. However, existing decision-making benchmarks rely on two simple assumptions: they formulate actions as selections from finite sets, and they assume unconditional environments in which every action is valid. These assumptions overlook the compositional nature and conditional feasibility of real-world actions. To address this limitation, we introduce \textsc{Condesion-Bench}, a benchmark for evaluating conditional decision-making in compositional action spaces. In \textsc{Condesion-Bench}, actions are defined as allocations to decision variables, and their validity is restricted by explicit conditions at the context, variable, and allocation levels. Through oracle-based evaluation of both decision quality and condition adherence, we provide a more realistic assessment of LLMs as decision-support tools.
\end{abstract}

% \begin{abstract}
% Large language models have been widely explored as decision-support tools in high-stakes domains by leveraging contextual understanding and reasoning capabilities. However, existing decision-making benchmarks suffer from two key limitations: they formulate actions as selections from finite sets, and they assume unconditional environments in which all actions are always valid. These assumptions fail to capture the compositional structure of real-world actions and the conditions that govern their feasibility. To address this gap, we introduce \textsc{Condesion-Bench}, a benchmark for evaluating conditional decision-making in compositional action spaces. In \textsc{Condesion-Bench}, actions are defined as allocations to decision variables, and their validity is restricted by explicit conditions at the context, variable, and allocation levels. Instantiated in the financial domain, \textsc{Condesion-Bench} enables oracle-based evaluation of both decision quality and condition adherence, providing a more realistic assessment of LLMs as decision-support tools.
% \end{abstract}

\section{Introduction}

Decision-making is the process of selecting an action to achieve a desired objective within a given context. To make sound decisions, decision-makers should understand complex contexts, compare multiple alternatives, and select the action with the best expected outcome~\citep{obi2017effective}. The combination of diverse information, numerous variables, and cognitive pressure complicates decision-making, even for humans~\citep{simon1977new, eigner2024determinants}. Recently, large language models (LLMs) have demonstrated promising capabilities as decision-support tools in high-stakes domains such as finance, healthcare, and supply chain management~\citep{li2025investorbench, hu2024mta, zhao2025aim, liudellma, chen2025decisionflow}.

\begin{figure}[t!]
    \centering
    \includegraphics[width=1\linewidth]{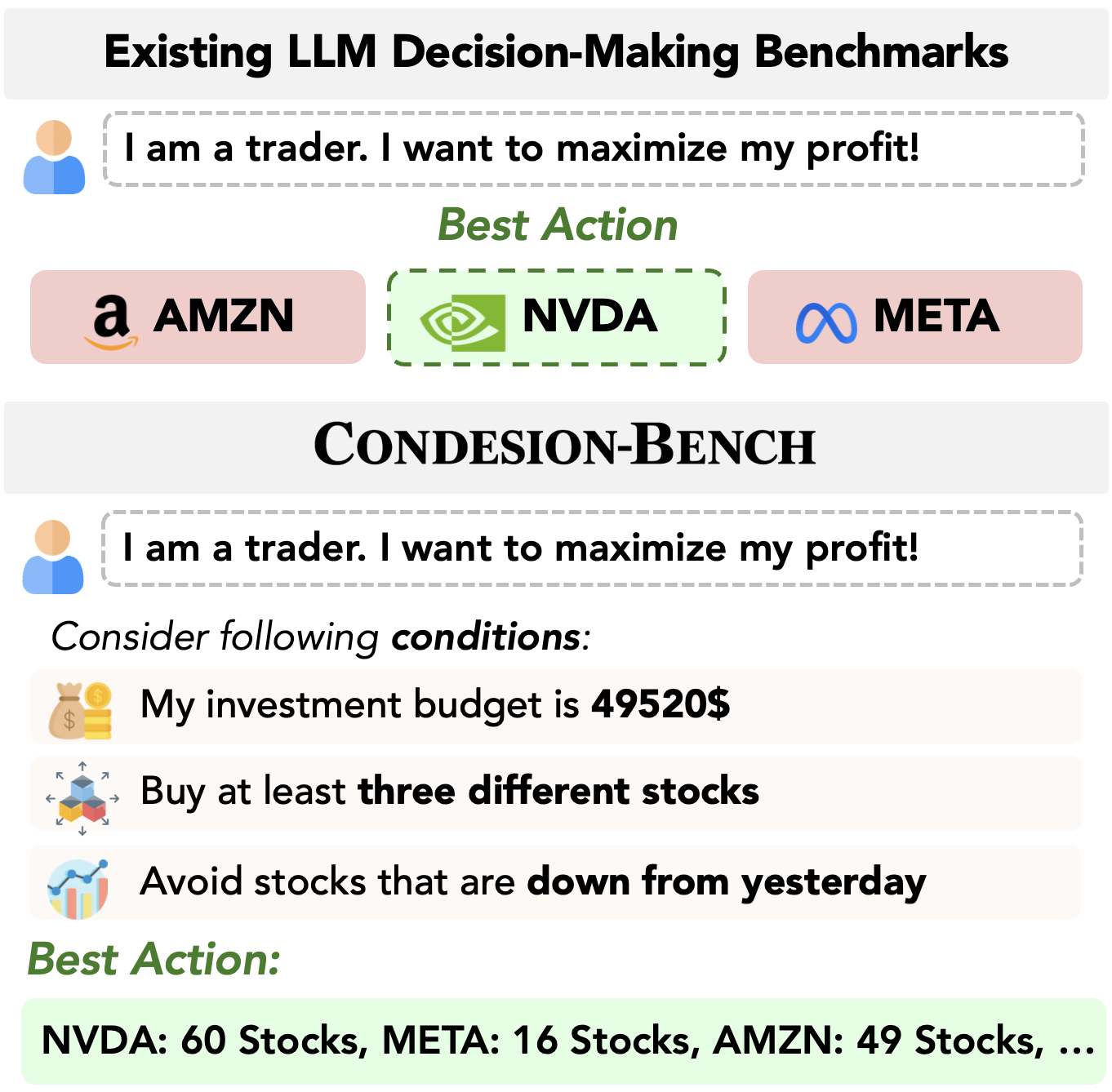}
    \caption{A comparison between conventional decision-making benchmarks (top) and \textsc{Condesion-Bench} (bottom). While prior benchmarks focus on selecting a single option from a fixed action set, conditional decision-making in compositional action space requires generating an action under explicit conditions.}
    \label{fig:figure_1}
    \vspace{-2.5em}

\end{figure}

However, as shown in Figure \ref{fig:figure_1}, existing benchmarks for decision-making rely on simplified problem formulations, resulting in two fundamental limitations. 
First, most benchmarks formulate decision-making as a \textbf{selection problem} over a finite set of pre-defined action candidates (e.g., buy NVDA, buy AMZN, or buy META). This formulation makes benchmark performance highly dependent on the construction of the candidate set~\citep{nguyen2024dynasaur, kim2025principles}. Moreover, in this formulation, models can appear effective by relying on shortcut reasoning that eliminates implausible candidates, thereby hindering accurate evaluation of LLM decision-making capabilities.
Second, existing benchmarks do not incorporate \textbf{explicit conditions} that restrict action feasibility into the decision-making process. Decisions are often conditioned by diverse factors such as available resources, legal regulations, organizational rules, or personal preferences~\citep{simon1972theories, march1996institutional, madanchian2025applications}. Under such factors, some actions in the action space can be infeasible, and effective decision-making therefore requires the ability to distinguish feasible actions from infeasible ones. Ignoring these conditions allows unrealistic or invalid actions to be considered feasible, leading to meaningless comparisons of actions and an overestimation of LLMs’ decision-making capabilities in real-world settings.

To address these limitations, we introduce \textsc{Condesion-Bench} (\textbf{Con}ditional \textbf{de}ci\textbf{sion}), a benchmark designed to evaluate conditional decision-making in compositional action space. Rather than relying on a fixed action set, we define a \textbf{compositional action space}, where an \emph{action} consists of selecting one or more \emph{decision variables} and assigning an \emph{allocation} to each selected variable (e.g., 100 units for A and 10 units for B). We further incorporate explicit conditions that determine action validity at three levels: (1) variable conditions, limiting which decision variables should be selected; (2) contextual conditions, which restrict actions based on the given context; and (3) allocation conditions, which govern the assignable amount of resources for decision variables. Together, these conditions enforce structured, realistic decision-making aligned with real-world requirements. Lastly, by using oracle-optimal actions as the evaluation reference, our benchmark enables accurate and interpretable assessment.

We systematically evaluate a wide range of LLMs, spanning from proprietary to open-source models, including models equipped with explicit reasoning capabilities on \textsc{Condesion-Bench}. Given the input context, objective, and conditions, LLMs generate the best action that can achieve the desired objective while satisfying explicit conditions. Our findings reveal that models largely differ in their capacity to comply with specified conditions. Notably, even models exhibiting high condition satisfaction struggle to generate actions to achieve the given objective. We also conduct additional analyses to uncover what current LLMs excel at and where they fail.

Our contributions are threefold:
\begin{itemize}
\itemsep0.05em 
    \item We reformulate decision-making with a compositional action space, where actions are allocations to decision variables rather than fixed atomic choices.
    \item We introduce \textsc{Condesion-Bench}, a benchmark that models realistic, multi-level conditions—at the variable, contextual, and allocation levels—that constrain valid action space.
    \item We propose an oracle-based evaluation framework grounded in real-world data, which evaluates both the relative quality of decisions and their adherence to imposed conditions.
\end{itemize}

\begin{figure*}[t]
    \centering
    \includegraphics[width=1\linewidth]{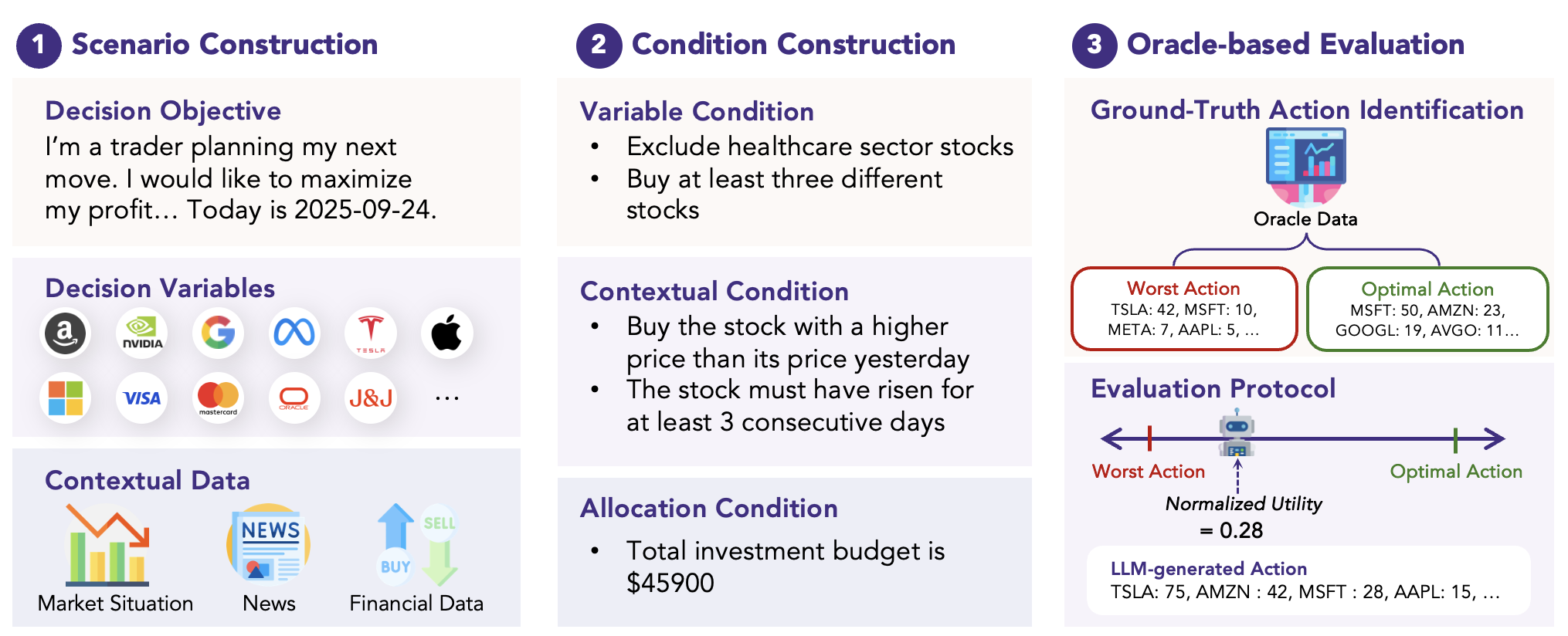}
    \caption{The overview of the dataset construction pipeline and the evaluation protocol of \textsc{Condesion-Bench}.}
    \label{fig:main_figure}
\end{figure*}
\section{Task Definition}
\label{definition}
Decision-making involves interpreting the given context, reasoning about the possible outcomes, and identifying the action that yields the highest utility (\ie the most desirable action for the decision-maker to achieve the objective)~\citep{fishburn1968utility, berger2013statistical}. Building on this definition, we propose conditional decision-making in compositional action space, which is a decision-making task where actions are generated compositionally and evaluated under explicit conditions, which we formalize in the following sections.

\subsection{Compositional Action Space}
In many decision-making benchmarks, actions are defined as a finite set of discrete options, and decision-making is formulated as selecting one option from this set. This works well when the set of possible actions is small and fixed. However, in domains where decisions can take many different forms, defining actions solely through predefined options becomes restrictive. For example, in stock trading, benchmarks often define an action as selecting a single asset, whereas real trading actions involve specifying quantities for multiple assets at the same time. Representing such actions as predefined options would require listing all possible combinations, which quickly becomes impractical as the action space grows.

We therefore consider a compositional action space $\mathcal{A}$, where each action is constructed with compositional units that bind a decision variable with its allocation. An action $a$ is defined as:
\begin{equation}
    a = \big[(v_1, \alpha_1), \ldots, (v_n, \alpha_n)\big],
\end{equation}
where each $v_i$ is a decision variable and $\alpha_i$ is its corresponding allocation. This formulation avoids restricting decisions to a predefined set of options, allowing actions to be specified directly.

Also, the allocation $\alpha_i$ can take different values depending on the type of the decision variable. Especially in the financial domain, decision variables are defined as buying particular stocks, and allocations are the amount of purchase. More generally, based on the decision variables, allocation values can be binary, integer, or continuous. This formulation allows decision variables of different types to be considered jointly within a single action and is applicable to a wide range of domains~\footnote{In medical triage, decision variables can simply indicate whether a patient receives a treatment or not, resulting in a binary allocation. In autonomous driving, decision variables correspond to control inputs such as speed, and the allocations are continuous values directly specifying the control signals.}.

% Also, the allocation $\alpha_i$ can take different values depending on the type of the decision variable. For example, in medical triage, a decision variable may represent whether a specific patient receives an intervention, where the corresponding allocation is binary, indicating a yes–no decision. In financial decision-making, a variable may correspond to a particular asset, and the allocation specifies a discrete quantity to purchase. In autonomous driving, decision variables may correspond to control components, and the allocations are continuous, representing real-valued control signals. This formulation allows decision variables of different types to be considered jointly within a single action.

\subsection{Conditional Decision-Making}
In real-world, human instructions include explicit conditions and require models to satisfy them~\citep{yang2023foundation}. Conditional decision-making evaluates whether a model can follow such condition-based instructions and produce decisions that remain aligned with the underlying objective. It consists of four main components: \textit{scenario}, \textit{condition}, \textit{action}, and \textit{utility}. The decision-maker aims to choose the action that yields the highest utility for a given scenario under the specified condition.

Let $\mathcal{S}$ denote a scenario, which represents a textual description of the current decision context such as objects, given variables, and contextual data that can affect decision making (\eg news data). Let $\mathcal{C} = \{c_1, \ldots, c_m\}$ be a set of conditions. In each scenario $\mathcal{S}$, a utility function $U_\mathcal{S}: \mathcal{A} \rightarrow \mathbb{R}$ assigns a scalar value to the action. The decision-maker’s objective is to identify an action that maximizes the utility while satisfying the given conditions. To achieve this objective, the decision-maker selects the action $a^*$, yielding the highest expected utility:
\begin{equation}
    a^* = \argmax_{a\in\mathcal{A}} \mathbb{E}\big[U_\mathcal{S}(a)\big] \quad \text{s.t. } \prod_{i=1}^m f(a, c_i)=1
\end{equation}
where $f(a, c_i)=1$ if an action $a$ satisfies condition $c_i$, and $f(a, c_i)=0$ otherwise.

\section{\textsc{Condesion-Bench}}
\label{method}

In this section, we introduce the dataset generation pipeline and our evaluation protocol for \textsc{Condesion-Bench}. As shown in Figure~\ref{fig:main_figure}, the pipeline consists of three stages. We design a scenario that defines the full context presented to the model (\S\ref{method:scenario}) and construct conditions that restrict the feasible action space within the given scenario (\S\ref{method:condition}). Finally, under each condition, we enumerate all possible actions and compute the best and worst actions using an oracle-based approach, which serves as the basis for our evaluation (\S\ref{method:evaluation}). % The fully automated pipeline enables scalable evaluation of both current and future models (see Appendix \ref{appendix:dataset} for details).

\subsection{Scenario Construction}
\label{method:scenario}
In \textsc{Condesion-Bench}, a scenario denotes a textual description of the current decision context faced by a decision-maker.
Each scenario consists of a decision objective, available decision variables, and a collection of contextual data, such as news or market situation relevant to the decision.

\paragraph{Decision Objective.}
We instantiate scenarios in the financial domain, where the objective is to maximize portfolio profit over a single trading day. Decision-makers invest at market open and liquidate all positions at market close, enabling explicit computation of utility. \textsc{Condesion-Bench} focuses on the financial domain as it provides diverse types of contextual information and supports clearly verifiable conditions. In addition, the financial domain allows explicit computation of utility, enabling quantitative evaluation of decision quality.

\paragraph{Decision Variables.}
We define the decision variables as 15 individual stocks from top-ranked S\&P 500 companies to avoid excessive market volatility and enable stable evaluation of decisions. The number of stocks is limited to 15 as a large number of decision variables can make it harder for models to consider all options, blurring whether performance differences reflect reasoning quality or context overload. Moreover, we select stocks across multiple industry sectors to incorporate sector-level conditions. Stocks from the same corporate group are excluded to preserve decision diversity.

% \footnote{\citealp{liudellma} employ seven fixed action set and \citealp{chen2025stockbench} employ 20 stocks for evaluation. While our benchmark can also be extended, we fix the number to 15 in order to reduce random guessing.}

% We define the decision variables as 15 individual stocks selected from the top-ranked companies in the S\&P 500, which ensures stable market behavior for reliable evaluation. We intentionally exclude companies that share the same corporate origin because such stocks often move together and reduce the diversity of decisions. We also select stocks across multiple industry sectors so that conditions can be defined over sector-level properties. This design allows us to construct a wide range of explicit and verifiable conditions, while keeping the action space interpretable within the context window of current LLMs.

\paragraph{Contextual Data.}
In each scenario $\mathcal{S}$, decision-makers seek to maximize end-of-market profit using structured contextual information available at the decision point. To enable informed decisions while avoiding excessive noise, we use three types of contextual data: historical market data, short-term news data, and overall market condition summaries. Specifically, daily price, volume data and quarterly financial statements are obtained from Yahoo Finance, sentiment-labeled news from the Massive API, and broad market summaries from Zacks. See Appendix~\ref{appendix:scenario} for details.

To prevent data leakage, all scenarios are constructed using only information available before the decision time, with no access to future outcomes. The scenarios span the period from \texttt{2024-10-01} to \texttt{2025-09-30}, which lies entirely beyond the stated knowledge cutoff of current models. 

\subsection{Condition Construction}
\label{method:condition}

\subsubsection{Definition}
The decision-making process is governed by three distinct types of conditions that determine how variables are filtered, evaluated, and allocated:

\textbf{Variable conditions} set the ground rules for the decision-making pool. They set hard boundaries on what can be chosen and how many should be selected, regardless of the current situation. For instance, a user can limit the types of options, like \textit{“excluding healthcare sector stocks”}, or set a group rule, like \textit{“buying at least three different stocks.”} This demands combinatorial reasoning over discrete variables—the ability to filter and identify valid options according to logical rules.

\textbf{Contextual conditions} further eliminate decision variables based on the current state of the environment. Unlike variable conditions, these are dynamic and depend on context-specific signals. For example, a user can ask to \textit{“buy the stock with a higher price than its price yesterday.”} This requires conditional reasoning—the ability to interpret context-dependent information and correctly evaluate the variables under the current state.

\textbf{Allocation conditions} constrain how much allocation can be assigned across the selected decision variables. They impose global limits on the total allocation, such as \textit{a user's budget} or \textit{capacity constraints}, and restrict the final action to remain feasible. This demands numerical reasoning—the ability to perform arithmetic operations, track cumulative quantities, and ensure global constraints are satisfied while optimizing for utility.

\subsubsection{Construction Process}

% \paragraph{Collection.}
For each condition type, we manually design fixed textual formats, resulting in 29 distinct templates in total, while the values within the formats are filled in randomly to avoid bias. These values include numbers, stock identifiers, or sector names. For example, a contextual condition may follow the format: \textit{“I should buy the stock that has achieved a volume increase for more than} \{\texttt{num}\} \textit{days”}, where \{\texttt{num}\} is randomly sampled.

For each scenario, one condition is randomly sampled from each condition type and applied jointly. If no feasible action exists under the sampled condition set, the conditions are discarded and regenerated for the same scenario. This process is repeated until feasibility is ensured.

Following this process, we construct 251 valid instances. This approach allows the benchmark to be easily scaled by generating additional instances as needed. Detailed examples for each condition are provided in Appendix \ref{appendix:condition}.

% \paragraph{Application to Scenarios.}
% For each scenario $\mathcal{S}$, we randomly assign one condition from each of the three types. We then check whether at least one feasible action exists under the assigned conditions. If no feasible action exists, we discard the condition set and regenerate a new one for the same scenario. We repeat this process until feasibility is ensured. Through this iterative procedure, we generate a total of 251 valid instances. This process allows us to generate additional instances as needed, enabling the benchmark to scale to larger datasets and broader evaluations.

\subsection{Oracle-based Evaluation}
\label{method:evaluation}
\subsubsection{Ground-Truth Action Identification}
In \textsc{Condesion-Bench}, utility is defined as the realized profit at the end of the market. A decision-maker generates an action that is likely to maximize end-of-market profit (\ie ground-truth utility), while satisfying all assigned conditions. However, accuracy-based evaluation, which is widely adopted in previous benchmarks, is not suitable because exact matching provides little insight into how close a generated action is to the optimum in a vast action space. For each instance, we therefore identify two reference actions using oracle information available after market close. % These reference actions are constructed differently depending on whether the evaluated action satisfies the given conditions or not.

Since the decision variables and their allocations are bounded by the imposed conditions, the resulting feasible action space is finite and explicitly defined. Under these constraints, we can exhaustively enumerate all condition-satisfying actions. Assuming oracle access to the close price, we evaluate the profit of each feasible action and identify the actions that maximize and minimize the profit by formulating the problem as a Knapsack problem \citep{pisinger1998knapsack}. These two actions serve as reference points, providing the upper and lower bounds of utility within the feasible decision space. A detailed algorithm is provided in Appendix \ref{appendix:gt_utility}.

% Also, we consider actions that violate one or more conditions. When conditions are not satisfied, the action space becomes infinite where all actions are executable, such as spending an excessive amount of money. In this case, defining a single reference action is impossible. Instead, we define reference actions at the level of decision variables rather than full actions. Specifically, we use the decision variable with the highest per-unit utility and the one with the lowest per-unit utility as reference actions. This design enables meaningful ground-truth comparison even when the generated action lies outside the feasible space.

\subsubsection{Evaluation Protocol}
\paragraph{Evaluation Setting.}
In \textsc{Condesion-Bench}, we treat an LLM as the decision-maker, where the models estimate expected utility. Each test instance, consisting of a market scenario and a set of assigned conditions, is provided to the model as an input prompt. The model is instructed to output an action in a structured JSON format, specifying a set of decision variables along with their corresponding allocation values, where each allocation is constrained to be non-negative. Finally, the LLMs output the action that satisfies the conditions and yields the highest utility. We provide a qualitative example in Appendix \ref{appendix:example}.

\paragraph{Evaluation Metrics.}
Given a generated action, we evaluate whether it satisfies all assigned conditions. We define Decision Satisfaction Rate (DSR) and Condition Satisfaction Rate (CSR) as follows:
\begin{equation}
    \text{DSR}=\prod_{i=1}^mf(a, c_i)
\end{equation}
\begin{equation}
    \text{CSR}=\dfrac{1}{m}\sum_{i=1}^mf(a, c_i)
\end{equation}
where DSR evaluates whether the generated action is fully feasible under all conditions, while CSR measures how well the model understands and applies each individual condition.

To evaluate the optimality toward the objective, we assess model performance by measuring where the utility lies relative to the reference actions. We define this Normalized Utility (NU):
\begin{equation} 
    \mathrm{NU} = \dfrac{U_\mathcal{S}(a) - U_\mathcal{S}(a_{\min})}{U_\mathcal{S}(a_{\max}) - U_\mathcal{S}(a_{\min})}
\end{equation}
where $a_{\max}$ indicates the action with the highest utility and $a_{\min}$ indicates the action with the lowest utility among the actions satisfying the conditions. With the reference actions defined above, when the generated action satisfies all conditions, we can measure the relative quality of the action in the feasible action space.
\section{Experiment}
\subsection{Experimental Setup}
To comprehensively evaluate model performance on \textsc{Condesion-Bench}, we conduct experiments across a wide range of LLMs, spanning both proprietary and open-source models, as well as models with and without explicit reasoning capabilities. Specifically, we evaluate non-reasoning models such as GPT-4.1, Claude, Gemini, Llama and Mistral, alongside reasoning models including GPT-5, OpenAI o-series, and GPT-oss. This diverse selection enables us to assess whether LLMs can generate compositional actions that both satisfy given conditions and maximize utility. To prevent potential data leakage, we restrict our main experiments to models whose knowledge cutoffs precede the time period covered by the data used in \textsc{Condesion-Bench}. Results for more recent models with later knowledge cutoffs are reported in Appendix \ref{appendix:additional}. Please note that while \textsc{Condesion-Bench} currently covers a fixed time span, it can be readily updated to newer time spans via our fully automated data generation pipeline.

Since generation tasks are inherently stochastic, we set the temperature to 0. Moreover, to reduce potential positional bias, we randomly shuffle the order of the stocks for each scenario. Additional experimental details are provided in Appendix \ref{appendix:setup}.

\subsection{Research Questions}
To systematically evaluate conditional decision-making in LLMs within a compositional action space, we analyze how well LLMs adhere to given conditions during decision-making and how it relates to utility through three research questions:

\paragraph{RQ 1: Can LLMs generate actions that satisfy explicit decision conditions?}
To evaluate whether LLMs can generate feasible actions under explicit conditions, we assess the extent to which model-generated actions satisfy the imposed conditions. Specifically, we measure feasibility using two complementary metrics: DSR and CSR. This dual-metric evaluation allows us to distinguish between actions that completely violate all imposed conditions and those that partially satisfy the given conditions, revealing whether models fail due to isolated condition violations or systematic difficulty in jointly satisfying multiple conditions.

\begin{table}[t]
\centering
\resizebox{0.35\textwidth}{!}{
\begin{tabular}{ccc}
    \toprule
    \textbf{Model} & \textbf{DSR} & \textbf{CSR} \\
    \hline
    \rowcolor{figuregray}
    \multicolumn{3}{c}{\textbf{Non-reasoning Models}} \\
    \multicolumn{1}{l}{GPT-4.1} & 0.5139 & 0.7968 \\
    \multicolumn{1}{l}{GPT-4.1-mini} & 0.5299 & 0.7968 \\
    \multicolumn{1}{l}{Claude-3.5-Sonnet} & \textbf{0.6375} & \textbf{0.8499} \\
    \multicolumn{1}{l}{Claude-3.5-Haiku} & 0.5737 & 0.8234 \\
    \multicolumn{1}{l}{Gemini-2.0-Flash} & 0.3227 & 0.6467 \\
    \multicolumn{1}{l}{Gemini-2.0-Flash-Lite} & 0.2869 & 0.6175 \\
    \multicolumn{1}{l}{Llama-3.3-70B} & 0.4622 & 0.7676 \\
    \multicolumn{1}{l}{Llama-3.1-8B} & 0.1076 & 0.4449 \\
    \multicolumn{1}{l}{Mistral-Large} & 0.5140 & 0.7703 \\
    \multicolumn{1}{l}{Mistral-Small} & 0.3307 & 0.6202 \\
    
    \rowcolor{figuregray}
    \multicolumn{3}{c}{\textbf{Reasoning Models}} \\
    \multicolumn{1}{l}{GPT-5} & 0.8924 & 0.9641 \\
    \multicolumn{1}{l}{GPT-5-mini} & \textbf{0.9124} & \textbf{0.9708} \\
    \multicolumn{1}{l}{o3} & 0.8884 & 0.9628 \\
    \multicolumn{1}{l}{o4-mini} & 0.8606 & 0.9509 \\
    \multicolumn{1}{l}{GPT-oss-120B} & 0.7530 & 0.8792 \\
    \multicolumn{1}{l}{GPT-oss-20B} & 0.7052 & 0.8858 \\
    \bottomrule
\end{tabular}
}
\caption{Comparison of Decision Satisfaction Rate (DSR) and Condition Satisfaction Rate (CSR) across various LLMs. Bold indicates the best performance among each type of models.}
\label{tab:rq1}
\end{table}

\paragraph{RQ 2: Are feasible actions also optimal?}
To examine whether feasible actions are also optimal, we restrict the analysis to model-generated actions that satisfy all given conditions and evaluate their decision quality relative to the oracle optimum. Optimality is measured using Normalized Utility, which enables a consistent comparison of decision quality across instances with different scales and conditions. By jointly considering feasibility and utility, this evaluation examines whether models can simultaneously satisfy conditions and optimize the objective, or whether a trade-off emerges between condition adherence and decision optimality.

\paragraph{RQ 3: Can LLMs maintain objective alignment despite condition violations?}
To further analyze model behavior beyond condition satisfaction, we conduct an additional error analysis on failed actions to better understand how models behave when conditions are violated. When conditions are violated, the action space becomes infinite, as all actions are executable, such as spending an excessive amount of money. In such cases, defining a single reference action is impossible. Instead, we measure return on investment (ROI, \ie profit-to-cost ratio) and leverage Normalized ROI (NR), which measures how close a generated action’s ROI is to that of references, as an auxiliary metric. Further details are provided in Appendix \ref{appendix:roi}.

% To further analyze model behavior beyond condition satisfaction, we examine whether models produce poor actions or generate actions close to the objective. When conditions are not satisfied, the action space becomes infinite where all actions are executable, such as spending an excessive amount of money. In this case, defining a reference action is impossible. Instead, we leverage an auxiliary metric: Normalized ROI (NR), which measures how close the generated action’s ROI is to the reference actions, on a 0–1 scale. We provide details in Appendix \ref{appendix:roi}.
\section{Results}
\label{result}

\begin{table}[t]
\centering
\resizebox{0.4\textwidth}{!}{
\begin{tabular}{cccc}
    \toprule
    \textbf{Models} & \textbf{V} & \textbf{C} & \textbf{A} \\
    \hline
    \rowcolor{figuregray}
    \multicolumn{4}{c}{\textbf{Non-reasoning Models}} \\
    \multicolumn{1}{l}{GPT-4.1-mini} & \textbf{91.63} & 76.10 & 71.31 \\
    \multicolumn{1}{l}{GPT-4.1} & 89.64 & 79.68 & 69.72 \\
    \multicolumn{1}{l}{Claude-3.5-Haiku} & 90.04 & 85.66 & 71.31 \\
    \multicolumn{1}{l}{Claude-3.5-Sonnet} & 90.44 & \textbf{93.63} & 70.92 \\
    \multicolumn{1}{l}{Gemini-2.0-Flash} & 90.84 & 77.29 & 63.35 \\
    \multicolumn{1}{l}{Gemini-2.0-Flash-Lite} & 88.45 & 60.16 & 45.42 \\
    \multicolumn{1}{l}{Llama-3.3-70B} & 88.05 & 72.51 & 69.72 \\
    \multicolumn{1}{l}{Llama-3.1-8B} & 56.97 & 46.61 & 29.88 \\
    \multicolumn{1}{l}{Mistral-Large} & 76.89 & 71.71 & \textbf{82.47} \\
    \multicolumn{1}{l}{Mistral-Small} & 74.50 & 66.53 & 45.02 \\
    
    \rowcolor{figuregray}
    \multicolumn{4}{c}{\textbf{Reasoning Models}} \\
    \multicolumn{1}{l}{GPT-5} & 92.43 & 96.81 & \textbf{100.00} \\
    \multicolumn{1}{l}{GPT-5-mini} & \textbf{94.02} & \textbf{97.21} & \textbf{100.00} \\
    \multicolumn{1}{l}{o3} & 92.83 & 96.41 & 99.60 \\
    \multicolumn{1}{l}{o4-mini} & 90.84 & 96.41 & 98.01 \\
    \multicolumn{1}{l}{GPT-oss-120B} & 90.04 & 86.06 & 87.65 \\
    \multicolumn{1}{l}{GPT-oss-20B} & 89.64 & 84.46 & 91.63 \\
    \bottomrule
\end{tabular}
}
\caption{Satisfaction rate (\%) for each condition type. V, C, A refer to variable, contextual, and allocation conditions, respectively. Bold indicates the best performance among each type of models.}
\label{tab:rq1_sub1}
\end{table}

\subsection{RQ 1: Can LLMs generate actions that satisfy explicit decision conditions?}
\label{result:rq1}
Table~\ref{tab:rq1} presents DSR and CSR across various LLMs on \textsc{Condesion-Bench}. It reveals a strong positive correlation between DSR and CSR, indicating that models effective at satisfying individual conditions tend to perform well under integrated conditions. A clear performance gap emerges between reasoning and non-reasoning models, with reasoning models consistently achieving substantially higher DSR and CSR. In contrast, non-reasoning models exhibit a large gap between DSR and CSR, indicating that they struggle to jointly handle multiple conditions even when they perform reasonably well on individual conditions.

\paragraph{Analysis on Type of Conditions.}

\begin{figure}[t]
    \centering
    \includegraphics[width=\linewidth]{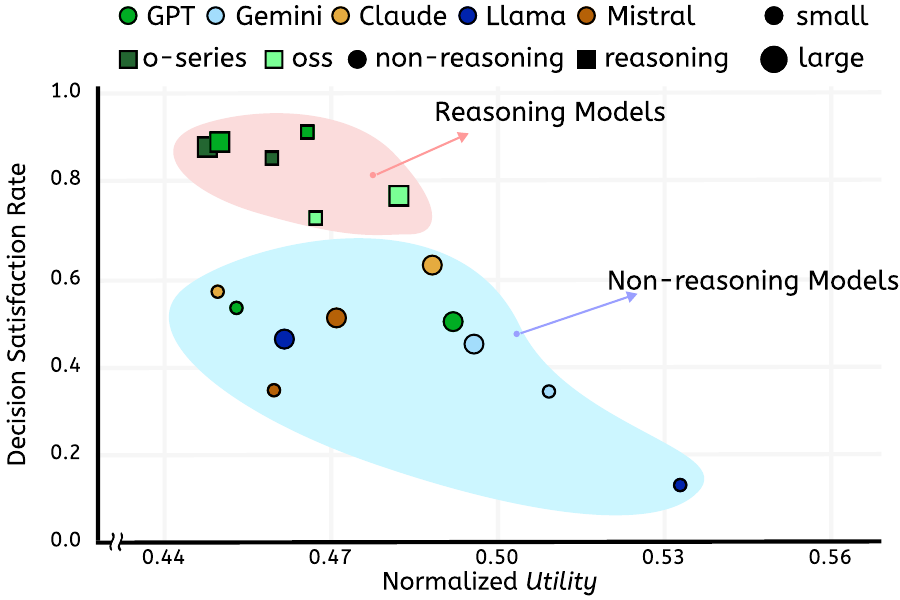}
    \caption{Decision Satisfaction Rate (DSR) over all generated actions and Normalized Utility (NU) computed on condition-satisfying actions.}
    \label{fig:rq2}
\end{figure}
In Table~\ref{tab:rq1_sub1}, we further analyze specific condition satisfaction rates by the type of conditions. Both reasoning and non-reasoning models exhibit comparable performance on variable conditions, whereas substantial gaps emerge for contextual and allocation conditions. This difference can be attributed to the reasoning complexity of each condition. Variable conditions can often be satisfied once the decision variable is identified. In contrast, contextual and allocation conditions demand multi-step reasoning grounded in the current scenario, such as identifying feasible decision variables from context or verifying allocation validity through cost calculations. These results suggest that the advantage of reasoning models becomes clearer when tasks require multi-step reasoning based on contextual information.

\subsection{RQ 2: Are feasible actions also optimal?}
\label{result:rq2}
To examine whether feasible actions are also optimal, we analyze the relationship between Decision Satisfaction Rate (DSR) across all actions and Normalized Utility (NU) of condition-satisfying actions. As shown in Figure \ref{fig:rq2}, reasoning and non-reasoning models exhibit contrasting behaviors with respect to optimality. Reasoning models with higher DSR tend to attain comparatively lower NU, indicating that condition satisfaction alone does not necessarily lead to high utility. In contrast, non-reasoning models achieve higher NU despite lower DSR, suggesting a tendency to trade off condition satisfaction for utility. While the ideal action corresponds to the upper-right region of the plot (high DSR and high NU), none of the evaluated models fully reach this region. These results indicate that reasoning capability is not by itself sufficient to consistently produce actions that jointly satisfy conditions and maximize utility.

\begin{figure}[t]
    \centering
    \includegraphics[width=0.9\linewidth]{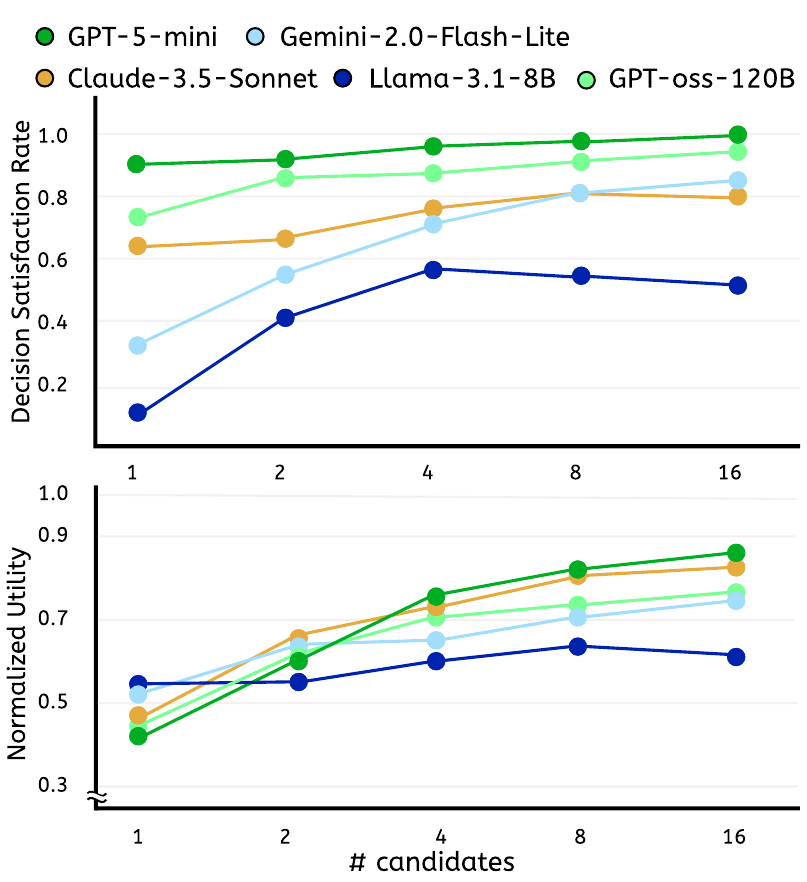}
    \caption{Comparison of performance regarding the number of generated action candidates. The upper plot indicates the proportion of cases in which at least one action satisfies all conditions, while the lower plot reports NU of the best action among the satisfying ones.}
    \label{fig:sampling}
\end{figure}

\paragraph{Impact of Sample Size on Action Quality.}
We examine whether models fail because they do not know how to construct utility-maximizing actions, or because they fail to generate such actions in a single attempt. To this end, instead of generating a single action, we instruct models to generate multiple action candidates per instance and analyze the resulting samples. 

Specifically, we address two questions: can the model generate feasible actions at all, and how good an action can it reach with sufficient sampling? We measure these using the probability of generating at least one feasible action and the normalized utility of the best feasible action among multiple candidates, evaluating five representative models from the upper-right region of Figure \ref{fig:rq2}.

As shown in Figure~\ref{fig:sampling}, when the number of samples increases, most models improve in both condition satisfaction and NU of the best action. This suggests that single-shot generation prioritizes feasibility over utility, while sufficient sampling enables higher-utility actions. However, some models still fail to produce a feasible action and achieve marginal improvement on NU, indicating that high NU scores for certain models (\eg Llama-3.1-8B) stem from few easy cases rather than consistent utility-maximizing behavior.

\subsection{RQ 3: Can LLMs maintain objective alignment despite condition violations?}
\label{result:rq3}

\begin{figure}[t]
    \centering
    \includegraphics[width=\linewidth]{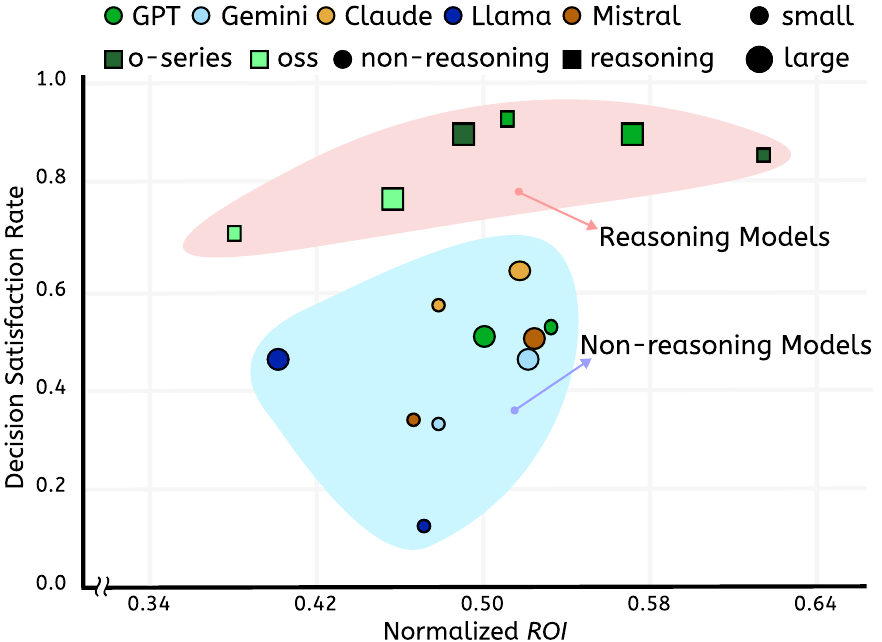}
    \caption{Decision Satisfaction Rate (DSR) over all generated actions and Normalized ROI (NR) computed on condition-violating actions.}
    \label{fig:rq3}
\end{figure}

We further analyze infeasible actions to examine whether condition-violating actions remain close to the objective. Figure \ref{fig:rq3} indicates NR of actions that violate conditions and reveals a clear distinction between reasoning and non-reasoning models. Reasoning models, while achieving high condition adherence, tend to generate profit-maximizing actions when conditions are not satisfied, indicating a balancing pattern between condition adherence and objective alignment. This behavior suggests that failures in reasoning models are not arbitrary but reflect a trade-off between competing objectives.

In contrast, non-reasoning models do not gain additional profit by violating conditions. Rather, they fall into a failure mode that is neither condition-satisfying nor profit-maximizing, and are outperformed by reasoning models even in terms of utility. Yet, since only actions satisfying conditions are treated as valid, strong performance requires models to achieve both high condition adherence and high utility, rather than excelling in only one.

\section{Related Work}

\paragraph{LLMs for Decision-Making.}
LLMs have emerged as a powerful tool for supporting decisions due to their capabilities to process long-context data and conduct logical reasoning toward action decisions~\citep{hu-etal-2025-define}. From low-stakes domains~\citep{liu2023g, nie2023importance, shinn2023reflexion, wang2025largelanguagemodelsoperations} to high-stakes domains~\citep{kwon2024large, maolanguage, yu2024fincon, simchi2026large, mainali2025classical, lu-etal-2025-strux}, recent works enhance decision-making with LLMs. DeLLMa~\citep{liudellma} enhances decision-making capabilities by predicting the future states and DecisionFlow~\citep{chen2025decisionflow} by equipping LLMs with structured reasoning. However, previous works define decision-making as a selection problem over a finite set of pre-defined options in unconditional environments. The simplified setting limits the reliable evaluation of readiness for deployment in real-world decision-making. In this work, we aim to evaluate LLM capabilities to generate compositional actions under explicit conditions.

\paragraph{Decision-Making in Constrained Spaces.}
Selecting actions in constrained spaces has been studied in various fields. Traditional reinforcement learning researches define this process as constrained Markov decision process (CMDP) and finding the optimal policy in CMDP as constrained optimization~\citep{altman1995constrained, achiam2017constrained, kotary2021end}. In this formulation, constraints restricting the feasible policies make the process of optimization more complicated. Several researches further adopt constrained optimization to solve the problems in practical scenarios~\citep{badanidiyuru2018bandits, wang2022learning, kachuee2023constrained, bermudez2023distributional, jiang2024followbench, fang2025offline}. These works recognize that constraints play a critical role in shaping feasible decisions and that ignoring them leads to unsafe or impractical outcomes. Building on prior works, our study advances the problem formulation to inference-time action generation under explicit conditions. In \textsc{Condesion-Bench}, the goal is not to learn or optimize a policy over repeated interactions, but to generate a valid and high-quality action in a single step under given conditions.

\section{Conclusion}
This paper introduces \textsc{Condesion-Bench}, a benchmark designed to evaluate the conditional decision-making of large language models within compositional action spaces. By explicitly modeling conditions and expanding the action space, our benchmark captures the feasibility structures inherent in real-world decision-making scenarios that are largely overlooked by existing benchmarks. Evaluations across a diverse set of models reveal that, although a number of LLMs can satisfy individual conditions, they struggle to jointly satisfy multiple conditions while producing high-quality decisions. These results reveal fundamental limitations in current models’ reasoning under conditions and compositional action spaces, and underscore the need for evaluation frameworks that jointly assess both decision feasibility and decision quality.

% This paper introduces \textsc{Condesion-Bench}, a benchmark designed to evaluate the conditional decision-making capabilities of large language models in compositional action spaces. Through explicit modeling of conditions and enlarged action space, we capture the feasibility structure inherent in real-world decision scenarios that are overlooked by prior benchmarks. Our evaluations across a diverse set of models show that while many LLMs can satisfy isolated conditions, jointly satisfying those conditions and generating high-quality decisions remains challenging. These findings reveal limitations in models’ reasoning under constrained action spaces and highlight the need for evaluation frameworks that jointly assess decision quality and feasibility. 
% We expect our novel benchmark to serve as a foundation for advancing LLMs toward more effective and reliable decision-support tools.
\section*{Limitations}
First, our proposed benchmark is currently restricted to the financial domain. We focus on the finance domain due to the restricted accessibility of real-world data in other high-stakes domains, where privacy, legal, and ethical constraints limit the availability of diverse scenarios and reliable reward signals. Although finance is a representative domain of high-stakes decision-making, model performance in conditional decision-making settings can differ across domains. Based on the three types of conditions defined in our work, future works can expand to other domains by designing domain-specific concrete conditions, decision variables and allocation schemes. Such extensions would enable broader evaluations of conditional decision-making across diverse real-world contexts.

Second, while we use three high-level types of conditions in our benchmark, each condition type can admit a richer and more complex set of concrete conditions depending on domain, circumstances, and personal preferences. Moreover, our current benchmark focuses on hard conditions that must be strictly satisfied. In practical decision-making, however, conditions may be soft—where violations are allowed at a cost—or implicit, reflecting unstated preferences or assumptions, which are not modeled in our current setting. Extending the framework to capture such richer conditions remains an important direction for future work.

Lastly, our benchmark focuses on daily trading scenarios within a limited time span, where decision-makers enter at market open and exit trades at market close. Therefore, the results may not generalize to longer-term investment settings, where holding trades across multiple days can lead to different outcomes. In addition, decision patterns of LLMs may vary when evaluated on data from different time spans or market periods, which are not covered in our current setup.

\bibliography{reference}

\newpage
\appendix

\section{Details on Dataset Generation}
\label{appendix:dataset}
\subsection{Scenario Construction}
\paragraph{Decision Objective.}
As we focus on the financial domain, the decision objective provided to LLMs is to maximize portfolio profit over a single trading day. Specifically, we use a text template \textit{"I'm a trader planning my next move. I would like to maximize my profit. Today is} \{\texttt{date}\}\textit{."} We construct one instance for each trading day in the period spanning \texttt{2024-10-01} to \texttt{2025-09-30}, resulting in 251 instances.

\begin{figure*}[ht]
    \centering
    \includegraphics[width=1\linewidth]{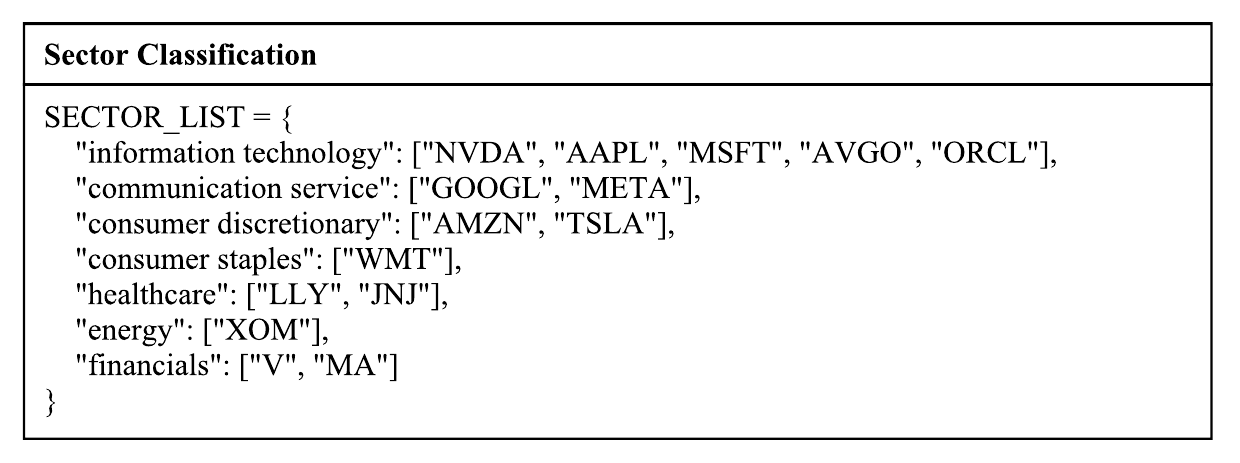}
    \caption{Sectors and following stock (decision variable) lists used for \textsc{Condesion-Bench}.}
    \label{fig:Sector Classification}
\end{figure*}

\paragraph{Decision Variables.}
In \textsc{Condesion-Bench}, we set the decision variables as fifteen individual stocks from top-ranked S\&P 500 companies, which are Nvidia, Apple, Microsoft, Amazon, Google, Broadcom, Meta, Tesla, Eli Lilly, Walmart, Visa, Oracle, ExxonMobil, Mastercard, and Johnson \& Johnson.
For each stock, we divide the sector according to the Global Industry Classification Standard (GICS), which is shown in Figure \ref{fig:Sector Classification}.

For prompts, we map the ticker and name of the company to normalize entity mentions (ticker \& company name) and improve consistency and readability in model inputs/outputs. We provide details in Figure \ref{fig: ticker-company name}.

\paragraph{Contextual Data.}
To enable informed decisions while avoiding excessive noise, we collect three types of contextual data:

\begin{itemize}[leftmargin=0.5em]
    \item \textbf{Historical Market Data:} 
    We collect daily price and trading volume data from the preceding two weeks, as well as the most recent quarterly balance sheets and income statements for each stock. These data are obtained from Yahoo Finance~\footnote{https://developer.yahoo.com/api/} and capture recent market dynamics and firm fundamentals.

    \item \textbf{Short-term News Sentiment Data:}
    We gather sentiment-labeled news articles published within the past three days from Massive~\footnote{https://massive.com/}, a stock-market data API. These signals reflect recent firm-specific and macroeconomic news that may influence short-term price movements.

    \item \textbf{Overall Market Condition Data:}
    We collect daily summaries of broad market conditions from Zacks~\footnote{https://www.zacks.com/stock/quote/API}, which are further summarized using \texttt{gpt-4o} to produce concise representations of prevailing market trends.
\end{itemize}

To prevent data leakage, all scenarios are constructed using only information available prior to the decision time, with no access to future outcomes. The scenarios span the period from \texttt{2024-10-01} to \texttt{2025-09-30}, which lies entirely beyond the stated knowledge cutoff or release date of current LLMs. 
\label{appendix:scenario}

\subsection{Condition Construction}
\label{appendix:condition}
As described in Section \ref{method}, we categorize conditions into three types. 

\textbf{Variable conditions} restrict the decision variables themselves: they may constrain the number of selected stocks, require/exclude specific tickers or particular sectors. These are expressed as conditions $\mathcal{C}$ using the natural-language forms in Table~\ref{condition_types}.

\textbf{Contextual conditions} are derived from the provided context—specifically, the price, volume, balance sheet, income statements, and news signals—and are also expressed to the model using the natural-language templates in Table \ref{condition_types}. These conditions can take diverse comparative forms (e.g., higher/lower, more/less than, increase/decrease), depending on the underlying context. 

Finally, \textbf{allocation conditions} impose resource limits on the action, most notably through a budget constraint on the total amount that can be invested. 

Given the natural-language condition set $\mathcal{C}$ together with the objective $\mathcal{O}$, the model is required to produce a decision (action) that satisfies all constraints while optimizing the objective.

\subsection{Reference Action Generation}
\label{appendix:gt_utility}

The pseudo algorithm for generating reference action is provided in Algorithm \ref{alg:reference_action_knapsack}. This algorithm represents a knapsack-based search that, for each possible cost level, maintains the action yielding the maximum accumulated profit or loss, and selects the resulting reference actions accordingly.

\section{Qualitative Examples}
\label{appendix:example}
\begin{figure*}[!t]
    \centering
    \includegraphics[width=1\linewidth]{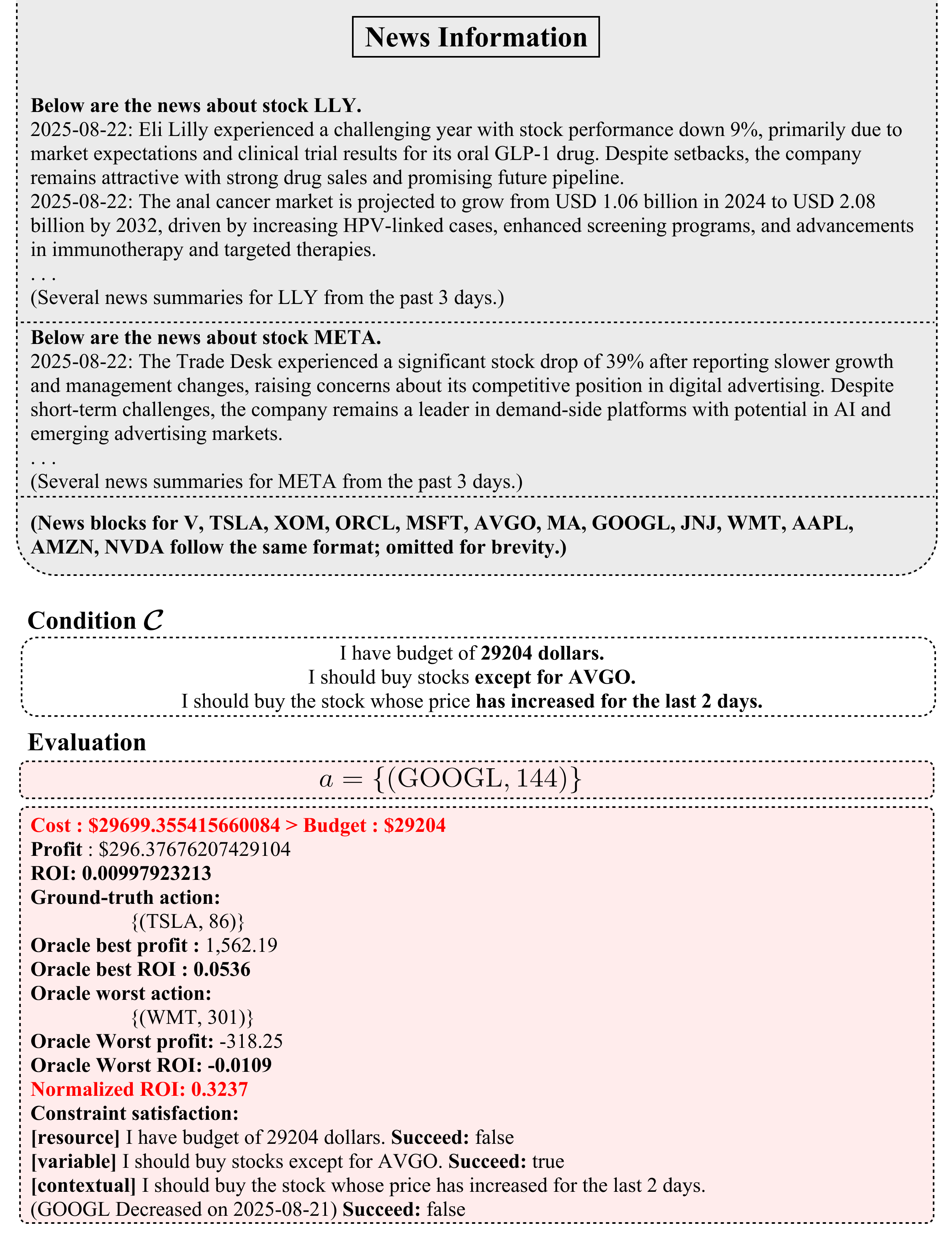}
    \caption{Qualitative example of scenario $\mathcal{S}$, condition $\mathcal{C}$, action $a$ (infeasible action) and evaluation on \textsc{Condesion-Bench}.}
    \label{fig:scenario2}
\end{figure*}

We present qualitative examples to illustrate how our benchmark evaluates the performance of LLMs. In particular, we show how the scenario $\mathcal{S}$ and conditions $\mathcal{C}$ are given and how we evaluate the generated action from LLMs. Figure~\ref{fig:scenario1} shows the evaluation of the action satisfying the conditions, and Figure~\ref{fig:scenario2} shows the violated action.

% As illustrated in Figure \ref{fig:main_figure}, \textsc{Condesion-Bench} constructs each evaluation instance by combining the scenario $\mathcal{S}$ and conditions $\mathcal{C}$. Using the benchmark’s oracle data, which in our setting corresponds to the same-day closing prices, we identify feasible actions that satisfy the conditions and select both the optimal action and the worst action within the feasible set. These two reference actions are then used to evaluate a model generated action $a$ via Normalized Utility, which measures the action’s relative utility with respect to the oracle best and oracle worst under the same conditions. A concrete example is provided in Table \ref{tab:qualitativeEx}.

\section{Additional Results}
\label{appendix:additional}
We conduct experiments to evaluate recent models with an up-to-date knowledge cutoff. Among 251 instances, we sample 126 instances ranging from \texttt{2025-04-01} to \texttt{2025-09-30} for additional results. Here, we use four metrics, SNU, FNU, DSR, and CSR, where SNU denotes NU of successful actions that satisfy conditions, and FNU denotes NU of failed actions. As shown in Table \ref{tab:additional}, reasoning models consistently outperform non-reasoning models in terms of constraint satisfaction, as reflected by substantially higher DSR and CSR. Interestingly, several recent models struggle more with condition satisfaction than previous ones, suggesting that improved general performance does not necessarily translate to better conditional decision-making. Conversely, we observe that NU improves for some recent models, even when their ability to satisfy conditions does not. Nevertheless, the overall patterns remain largely consistent with those observed in the main experiments.
\begin{table}[t]
    \centering
    \resizebox{\linewidth}{!}{
    \begin{tabular}{c c c c c}
    \toprule
    \textbf{Model}  &  \textbf{SNU} & \textbf{FNU}  & \textbf{DSR} & \textbf{CSR} \\
    \hline
    
    \rowcolor{figuregray}
    \multicolumn{5}{c}{\textbf{Non-reasoning Models}} \\
    \multicolumn{1}{l}{Nova-Pro}  & 0.4215 & 0.4797 & 0.3680 & 0.7173 \\
    \multicolumn{1}{l}{Nova-Flash}  & 0.4509 & 0.4621 & 0.2640 & 0.6347  \\
    
    \rowcolor{figuregray}
    \multicolumn{5}{c}{\textbf{Reasoning Models}} \\
    \multicolumn{1}{l}{Claude-4.5-Sonnet}  & 0.5048 & 0.4963 & 0.5968 & 0.8199 \\
    \multicolumn{1}{l}{Claude-4.5-Haiku}   & 0.4492 & 0.4468 & 0.5680 & 0.8427 \\
    \multicolumn{1}{l}{Grok-4}  & 0.4130 & 0.4693 & \textbf{0.8968} & \textbf{0.9656} \\
    \multicolumn{1}{l}{Grok-4-Fast}  & 0.3796 & 0.5126 & 0.8400 & 0.9440 \\
    \multicolumn{1}{l}{Gemini-2.5-Pro}  & 0.4587 & 0.4433 & 0.8880 & 0.9627 \\
    \multicolumn{1}{l}{Gemini-2.5-Flash}  & \textbf{0.5320} & \textbf{0.5267} & 0.4320 & 0.7093 \\
    \bottomrule
    
    \end{tabular}
    }
    \caption{Additional results on \textsc{Condesion-Bench} conducted with zero-shot prompting. SNU denotes NU of successful actions that satisfy conditions and FNU denotes NU of failed actions.}
    \label{tab:additional}
\end{table}

\section{Implementation Details}
\subsection{Detailed Experiment Setups}
\label{appendix:setup}
We provide detailed implementation details for the models employed in Table \ref{tab:model_specification}. Proprietary models from OpenAI are inferred using OpenAI API~\footnote{https://platform.openai.com/docs/models}. Models from the other providers are inferred via OpenRouter API~\footnote{https://openrouter.ai/models}. Also, we leverage default parameters except for temperature (set to 0) and response format (JSON): reasoning efforts to medium (only for reasoning models), top $p$ to 1, and $n$ to 1.

\subsection{Definition of Normalized Return}
\label{appendix:roi}
For actions that violate given conditions, we evaluate performance using return-on-investment (ROI, \ie profit-to-cost ratio) rather than absolute utility. Once conditions are not enforced, the action space becomes effectively unbounded, as models can arbitrarily scale investment amounts or select infeasible decision variables. In this setting, direct one-to-one comparisons between actions are no longer meaningful. For example, an action yielding \$10 profit from a \$1,000 investment is strictly worse than one yielding \$100 profit from the same cost; however, by scaling the former to a \$10,000 investment, both actions would appear equivalent under absolute profit, despite their fundamentally different quality.

To address this issue, we focus on whether a model selects actions that maximize utility per unit cost, independent of the total amount invested or feasibility constraints. Specifically, we compute per-unit profit, which reflects how efficiently an action converts cost into utility. The best action $a_{\max}$ is defined as investing in the option with the highest per-unit profit, while the worst action $a_{\min}$ corresponds to the lowest. Model-generated actions are evaluated based on their per-unit profit relative to this range, and the resulting score is normalized to obtain the Normalized Return. This metric allows us to consistently assess utility-maximization behavior even when conditions are violated and the action space becomes unconstrained. We define NR as follows:
\begin{equation} 
    \mathrm{NR} = \dfrac{R_\mathcal{S}(a) - R_\mathcal{S}(a_{\min})}{R_\mathcal{S}(a_{\max}) - R_\mathcal{S}(a_{\min})}
\end{equation}
where $R_\mathcal{S}(a)$ denotes ROI of action $a$ under scenario $\mathcal{S}$.

\subsection{Prompts for \textsc{Condesion-Bench}}

This section reports major prompts used in \textsc{Condesion-Bench}. 

\begin{figure*}[t]
    \centering
    \includegraphics[width=1\linewidth]{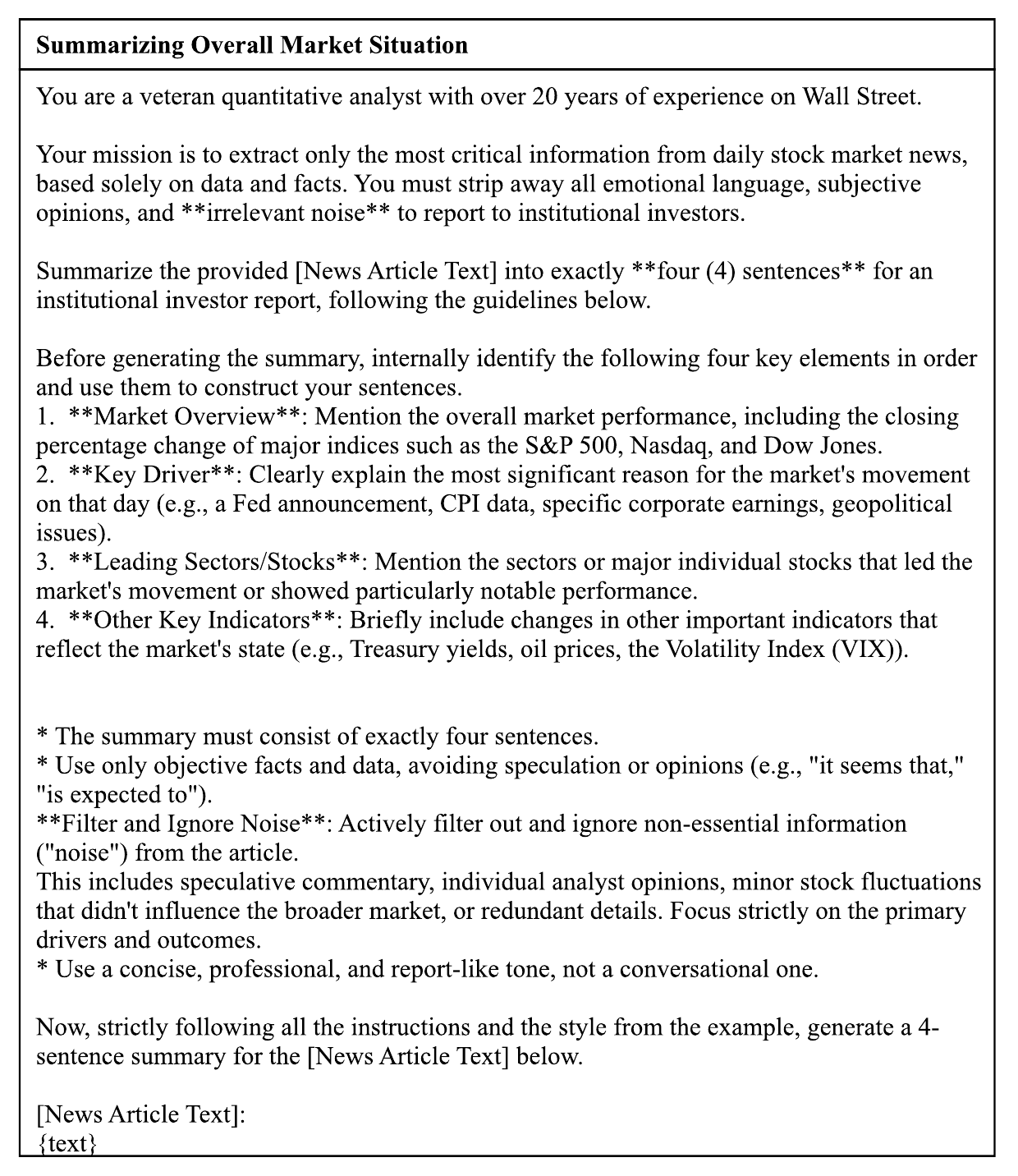}
    \caption{Prompt for summarizing overall market situation.}
    \label{fig:summarizing prompt}
\end{figure*}

\paragraph{Summarizing Overall Market Situation.}
We prompt the model to transform raw daily market news into a concise market-wide context summary. We enforce an institutional, objective tone and require the model to produce exactly four sentences, each corresponding to a predefined element: (1) market overview (major index moves), (2) key driver, (3) leading sectors/stocks, and (4) other key indicators (\eg yields, oil, VIX). The prompt for summarizing overall market situation is provided in Figure \ref{fig:summarizing prompt}.

\paragraph{Action Generation.}
We prompt the model to produce the final trading decision for each instance. We instruct an LLM to act as a rational decision-maker and select exactly one best action. The output is constrained to a strict JSON format with action instances containing \{\texttt{stock}\_\texttt{name}, \texttt{quantity}\} pairs; if no profitable action is identified, the model must return an empty list. We design the prompt by referring to zero-shot prompts used in \citet{liudellma}. The prompt for generating action is provided Figure \ref{fig:action-generation}.

\paragraph{Action Sampling.}
We also provide the prompt for experiments in Figure \ref{fig:sampling}. The only difference is the number of generated action candidates. In this case, the output is constrained to a strict JSON format containing a list of action instances. (\ie \{\texttt{index}, \texttt{instance}: [\texttt{stock}\_\texttt{name}, \texttt{quantity}]\}. The prompt for sampling actions is provided in Figure \ref{fig:sampling_prompt}.

% --- Appendix A_2 ---
\begin{table*}[t]
    \centering
    \small
    \begin{tabularx}{\textwidth}{@{} l X @{}}
        \toprule
        \textbf{Condition Type} & \textbf{Natural Language Form} \\
        \midrule
        % --- Variable ---
        \multicolumn{2}{l}{\textit{\textbf{1. Variable Conditions}}} \\
        \multirow{2}{*}{\hspace{1em}Individual} &
        I should buy stocks including \textbf{\{stock\}}.\\
        & I should buy stocks except for \textbf{\{stock\}}.\\
        \\
        \multirow{2}{*}{\hspace{1em}Number} &
        I should buy more than \textbf{\{num\}} kinds of stocks.\\
        & I should buy less than \textbf{\{num\}} kinds of stocks.\\
        \\
        \multirow{2}{*}{\hspace{1em}Sector} &
        I should only buy the stock of the company regarding   \textbf{\{sector\}}.\\
        & I should not buy the stock of the company regarding \textbf{\{sector\}}.
        \\
        \addlinespace
        \midrule
        % --- Contextual ---
        \multicolumn{2}{l}{\textit{\textbf{2. Contextual Conditions}}} \\
        \hspace{1em}\multirow{8}{*}{Price}
        & I should buy the stock with higher current price than the last price.\\
        & I should buy the stock with lower current price than the last price.\\
        & I should buy the stock that has achieved price increase more than \textbf{\{num\}} days for 2 weeks.\\
        & I should buy the stock that has achieved price decrease more than \textbf{\{num\}} days for 2 weeks.\\
        & I should buy the stock whose price has decreased for the last \textbf{\{num\}} days.\\
        & I should buy the stock whose price has increased for the last \textbf{\{num\}} days.\\
        & I should buy the stock whose current price is higher than the mean of last two weeks.\\
        & I should buy the stock whose current price is lower than the mean of last two weeks.\\
        \\
        \multirow{6}{*}{\hspace{1em}Volume} &
        I should buy the stock that has achieved volume increase more than \textbf{\{num\}} days for 2 weeks.\\
        & I should buy the stock that has achieved volume decrease more than \textbf{\{num\}} days for 2 weeks.\\
        & I should buy the stock whose volume has decreased for the last \textbf{\{num\}} days.\\
        & I should buy the stock whose volume has increased for the last \textbf{\{num\}} days.\\
        & I should buy the stock whose last volume is more than the mean of last two weeks.\\
        & I should buy the stock whose last volume is less than the mean of last two weeks.\\
        \\
        \multirow{2}{*}{\hspace{1em}Balance} &
        I should buy the stock of the company with more equity than liability.\\
        & I should buy the stock of the company with less equity than liability.\\
        \\
        % Income (4)
        \multirow{4}{*}{\hspace{1em}Income} &
        I should buy the stock of the company whose net income is more than \textbf{\{num\}} percent of their expenses.\\
        & I should buy the stock of the company whose net income is less than \textbf{\{num\}} percent of their expenses.\\
        & I should buy the stock of the company whose net income is more than \textbf{\{num\}} percent of their revenues.\\
        & I should buy the stock of the company whose net income is less than \textbf{\{num\}} percent of their revenues.\\
        \\
        % News (2)
        \multirow{2}{*}{\hspace{1em}News} &
        I should buy the stock without any negative news.\\
        & I should buy the stock with positive news.\\
        \addlinespace % 섹션 구분 여백
        \midrule
        
        % --- Allocation ---
        \multicolumn{2}{l}{\textit{\textbf{3. Allocation Conditions}}} \\
        \hspace{1em} Resource & I should buy more than \textbf{\{num\}} shares of any single stock. \\
        
        \bottomrule
    \end{tabularx}

    \caption{Types and manually generated natural language formats of conditions $\mathcal{C}$. }
    \label{condition_types}
\end{table*}

% --- Appendix A_3 ---
\begin{algorithm*}[t]
\caption{Algorithm for generating reference actions.}
\label{alg:reference_action_knapsack}
\small
\hrule
\vspace{0.3em}

\textbf{Notation:}   
$p_i^{open}, p_i^{close}$: open price and close price of $v_i$;
$w_i$: unit cost of $v_i$;  
$\Delta_i^{+}$, $\Delta_i^{-}$: unit profit/loss of $v_i$;  
$B$: budget;  
$dp^{+}[b], dp^{-}[b]$: maximum accumulated profit / loss with total cost $b$;
$\texttt{choice}^{+}[b], \texttt{choice}^{-}[b]$:stored actions achieving the maximum accumulated profit / loss at cost level $b$;
$O$: Oracle information

\vspace{0.3em}
\hrule
\vspace{0.5em}

\textbf{Input:} $\mathcal{S}$, $\mathcal{C}$, $B$, $O$

\textbf{Output:}  
Profit reference action $a_{\max}$, Loss reference action $a_{min}$ \hfill \textcolor{blue}{\texttt{\textbf{// this is a comment}}}

\vspace{0.5em}

\begin{algorithmic}[1]
\STATE \textcolor{blue}{\texttt{\textbf{/* Extract the per unit cost and profit/loss for each stock.}}} \hfill \textcolor{blue}{\texttt{\textbf{*/}}}
\STATE Extract $\mathcal{V}, p_i^{open}$ from $\mathcal{S}$
\STATE Extract $p_i^{close}$ from $O$
\FOR{each $v_i\in\mathcal{V}$}
    \STATE $w_i \leftarrow p_i^{open}$
    \STATE $\Delta_i^{+} \leftarrow p_i^{close}-p_i^{open}$
    \STATE $\Delta_i^{-} \leftarrow p_i^{open}-p_i^{close}$
\ENDFOR

\vspace{0.3em}
\STATE Initialize $dp^{+}[0]\leftarrow 0$, $dp^{-}[0]\leftarrow 0$
\STATE Initialize $dp^{+}[b],dp^{-}[b]\leftarrow -\infty$ for $b=1,\dots,B$
\STATE Initialize $\texttt{choice}^{+}[b],\texttt{choice}^{-}[b]\leftarrow\emptyset$

\STATE \textcolor{blue}{\texttt{\textbf{/* Traverse all cost levels.}}} \hfill \textcolor{blue}{\texttt{\textbf{*/}}}
\vspace{0.3em}
\FOR{$b=1$ \textbf{to} $B$}
    \FOR{each $v_i\in\mathcal{V}$}
        \IF{$b-w_i \ge 0$}
            \STATE \textcolor{blue}{\texttt{\textbf{/* Generate candidate action by increasing the allocation to $v_i$.}}} \hfill \textcolor{blue}{\texttt{\textbf{*/}}}
            \STATE \textcolor{blue}{\texttt{\textbf{/* Get $\alpha_i$ within the cost level $b-w_i$ and increase the allocation}}} \hfill \textcolor{blue}{\texttt{\textbf{*/}}}
            \STATE $\hat{a} \leftarrow \texttt{choice}[b-w_i]$
            \STATE $\hat{a} \leftarrow [(v_j,\alpha_j)]_{j=1}^{n} \text{ where } \alpha_i \leftarrow \alpha_i + 1$

            \STATE \textcolor{blue}{\texttt{\textbf{/* If candidate action satisfies the condition.}}} \hfill \textcolor{blue}{\texttt{\textbf{*/}}}
            \IF{$\prod_{k=1}^{m} f(\hat{a},c_k)=1$}
                \STATE \textcolor{blue}{\texttt{\textbf{/* If adding an allocation improves the maximum profit or loss at the corresponding cost level, the allocation is updated.}}} \hfill \textcolor{blue}{\texttt{\textbf{*/}}}
                \IF{$dp^{+}[b-w_i]+\Delta_i^{+} > dp^{+}[b]$}
                    \STATE $dp^{+}[b]\leftarrow dp^{+}[b-w_i]+\Delta_i^{+}$
                    \STATE $\texttt{choice}^{+}[b]\leftarrow \hat{a}$
                \ENDIF
                \IF{$dp^{-}[b-w_i]+\Delta_i^{-} > dp^{-}[b]$}
                    \STATE $dp^{-}[b]\leftarrow dp^{-}[b-w_i]+\Delta_i^{-}$
                    \STATE $\texttt{choice}^{-}[b]\leftarrow \hat{a}$
                \ENDIF
            \ENDIF
        \ENDIF
    \ENDFOR
\ENDFOR

\vspace{0.3em}

\STATE \textcolor{blue}{\texttt{\textbf{/* Find out the cost level with the highest profit and loss, respectively.}}} \hfill \textcolor{blue}{\texttt{\textbf{*/}}}
\STATE $b^{*}_{+}\leftarrow \arg\max_{0\le b\le B} dp^{+}[b]$
\STATE $b^{*}_{-}\leftarrow \arg\max_{0\le b\le B} dp^{-}[b]$

\STATE \textcolor{blue}{\texttt{\textbf{/* Assign the stored action at $b^{*}_{+}$ and $b^{*}_{-}$.}}} \hfill \textcolor{blue}{\texttt{\textbf{*/}}}
\STATE $a_{\max}\leftarrow \texttt{choice}^{+}[b^{*}_{+}]$
\STATE $a_{\min}\leftarrow \texttt{choice}^{-}[b^{*}_{-}]$

\STATE \textbf{return} $(a_{\max},a_{\min})$
\end{algorithmic}

\vspace{0.3em}
\hrule
\end{algorithm*}

% --- Appendix A_4 ---
\begin{figure*}[t]
    \centering
    \includegraphics[width=1\linewidth]{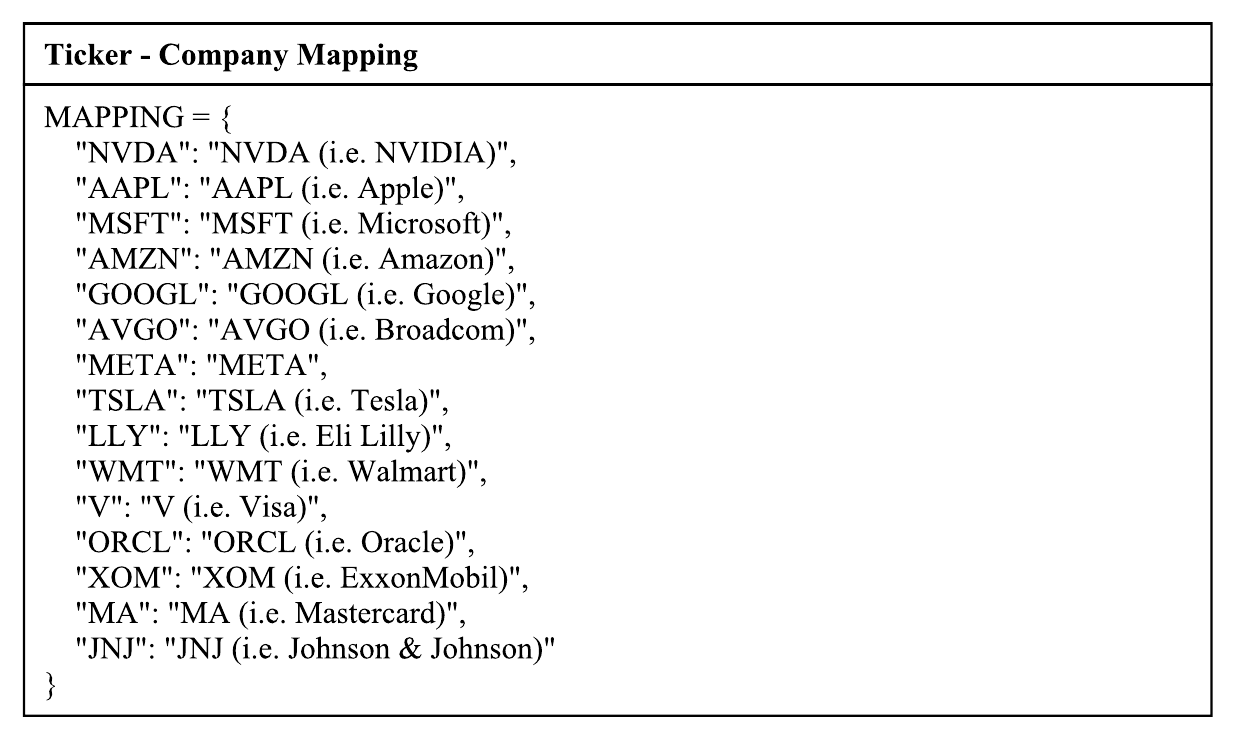}
    \caption{Mapping of stock tickers to company names used for \textsc{Condesion-Bench} construction.}
    \label{fig: ticker-company name}
\end{figure*}

%--- Appendix B_1 ---
\begin{figure*}[t]
    \centering
    \includegraphics[width=1\linewidth]{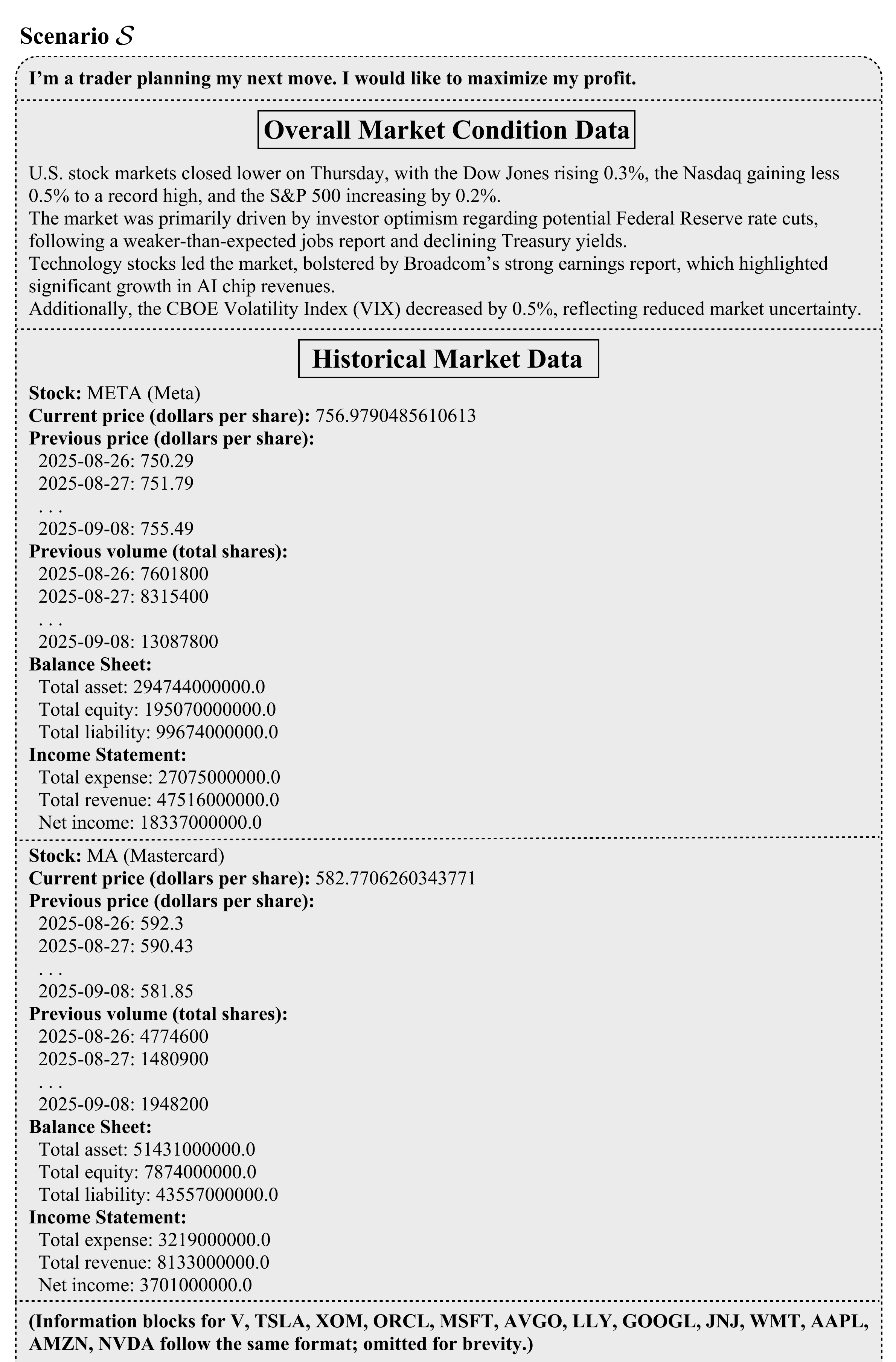}
    \addtocounter{figure}{0}
\end{figure*}

\begin{figure*}[!t]
    \centering
    \includegraphics[width=1\linewidth]{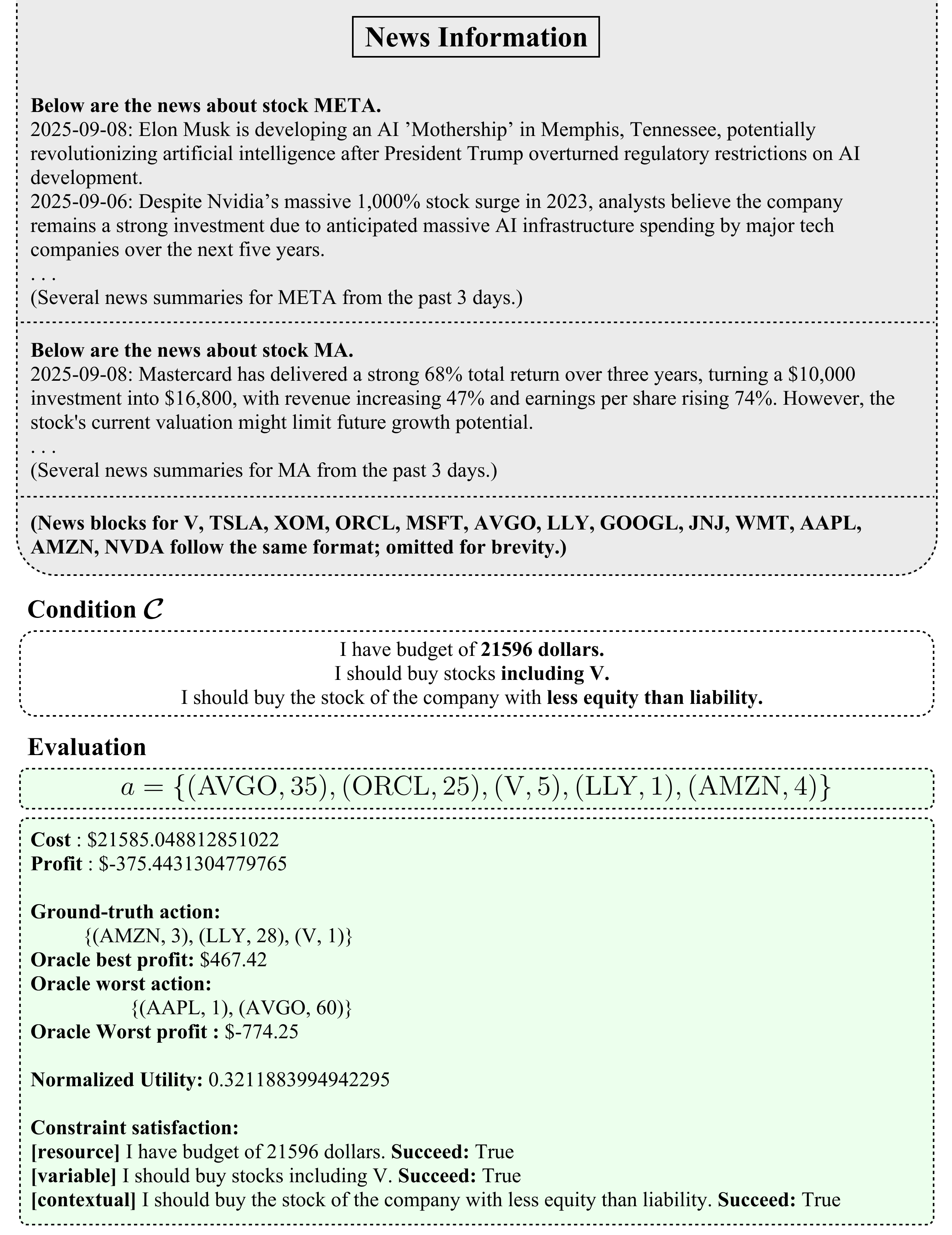}
    \caption{Qualitative example of scenario $\mathcal{S}$, condition $\mathcal{C}$, action $a$ (feasible action) and evaluation on \textsc{Condesion-Bench}.}
    \label{fig:scenario1}
\end{figure*}

\begin{figure*}[t]
    \centering
    \includegraphics[width=1\linewidth]{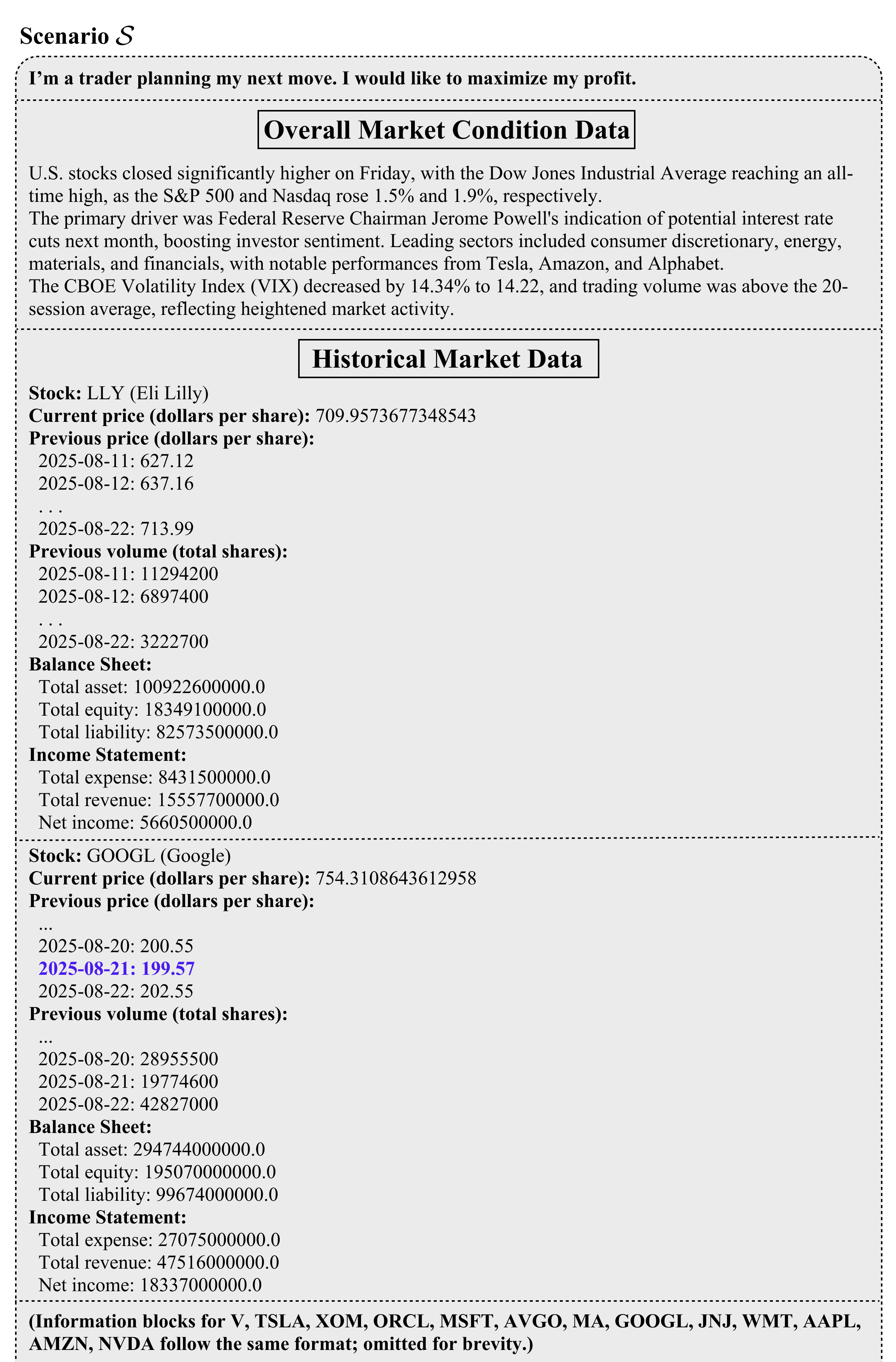}
    \addtocounter{figure}{0}
\end{figure*}

% --- Appendix D_1 ---
\begin{table*}[t]
\centering
\small
\setlength{\tabcolsep}{6pt}
\renewcommand{\arraystretch}{1.15}
\begin{tabular}{c c c c c c}
\toprule
\textbf{Reasoning} & \textbf{Type} & \textbf{Model} & \textbf{Source} & \textbf{Knowledge cutoff} & \textbf{Size}\\
\midrule

\multirow{12}{*}{X}
& \multirow{8}{*}{Proprietary} & gpt-4.1-2025-04-14             & \citep{gpt-4.1} & 2024.06 & - \\
&                              & gpt-4.1-mini-2025-04-14         & \citep{gpt-4.1} & 2024.06 & - \\
&                              & Claude-3.5-Sonnet-2024-10-22    & \citep{anthropic2024claude35sonnet_addendum}  & 2024.04 & - \\
&                              & Claude-3.5-Haiku-2024-10-22     & \citep{anthropic2024claude35_addendum_oct}  & 2024.07 & - \\
&                              & 	
gemini-2.0-flash-001     & \citep{geminiteam2023gemini}  & 2024.06 & - \\
&                              & gemini-2.0-flash-lite-001& \citep{geminiteam2023gemini}  & 2024.06 & - \\
&                              & Nova-Pro   & \citep{Intelligence2024}  & 2024.12$^*$ & - \\
&                              & Nova-Flash     & \citep{Intelligence2024}  & 2024.12$^*$ & - \\
\cline{2-6}
& \multirow{4}{*}{Open-source} & Llama-3.3-70B-Instruct        & \citep{metaai2024a_llama33_modelcard}  & 2023.12 & 70B\\
&                              & Llama-3.1-8B-Instruct         & \citep{grattafiori2024llama3herd}  & 2023.12 & 8B\\
&                              & Mistral-Large-2-Instruct-2407   & \citep{mistralai2024b_mistral_large_instruct_2411}  & 2023.10 & 123B \\
&                              & Mistral-Small-3.2-Instruct-2506       &  \citep{mistralai2024c_mistral_small_24b_instruct} & 2023.10 & 24B\\
\midrule

\multirow{12}{*}{O}
& \multirow{12}{*}{Proprietary} & gpt-5-2025-08-07     & \citep{openai2025gpt5systemcard}  & 2024.10 & - \\
&                              & gpt-5-mini-2025-08-07           & \citep{openai2025gpt5systemcard}  & 2024.05 & - \\
&                              & o3-2025-04-16                  & \citep{openai2025o3o4mini_systemcard} & 2024.06 & - \\
&                              & o4-mini-2025-04-16              & \citep{openai2025o3o4mini_systemcard} & 2024.06 & - \\
&                              & Claude-4.5-Sonnet              & \citep{anthropic_claude_sonnet_4_5_2024} & 2025.01 & - \\
&                              & Claude-4.5-Haiku              & \citep{anthropic_haiku_4_5_2025} & 2025.02 & - \\
&                              & Grok-4              & \citep{grok4} & 2024.11 & - \\
&                              & Grok-4-Fast              & \citep{grok4fast} & 2024.11 & - \\
&                              & Gemini-2.5-Pro              & \citep{comanici2025gemini25pushingfrontier} & 2025.01 & - \\
&                              & Gemini-2.5-Flash              & \citep{comanici2025gemini25pushingfrontier} & 2025.01 & - \\
\cline{2-6}
& \multirow{2}{*}{Open-source} & gpt-oss-120B         & \citep{openai2025gptoss_modelcard}  & 2024.06 & 120B \\
&                              & gpt-oss-20B          & \citep{openai2025gptoss_modelcard}  & 2024.06 & 20B \\
\bottomrule
\end{tabular}
\caption{Details on the models used for our experiments. For models that do not explicitly disclose a knowledge cutoff, the release date is used as a proxy and is marked with *.}
\label{tab:model_specification}
\end{table*}

% --- Appendix D_2 ---

\begin{figure*}[t]
    \centering
    \includegraphics[width=1\linewidth]{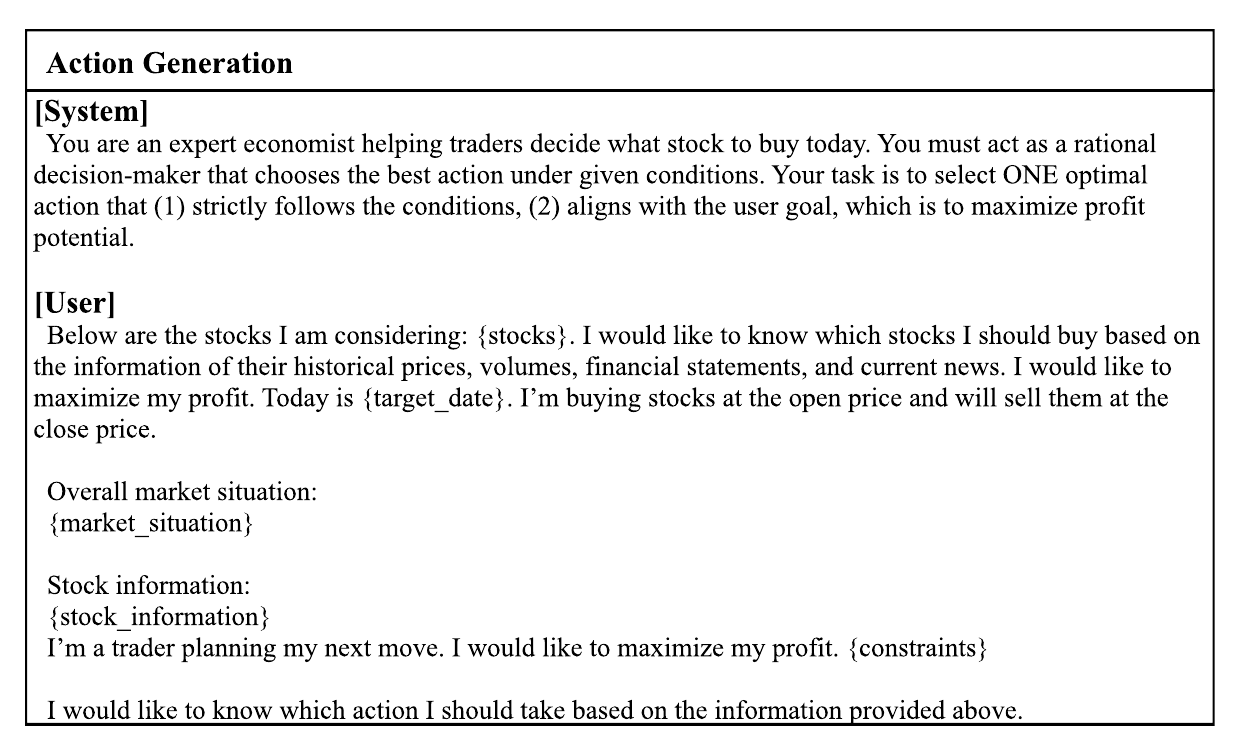}
    \caption{Prompt for generating final action. Input data are given in Figure \ref{fig:scenario1} and \ref{fig:scenario2}.}
    \label{fig:action-generation}
\end{figure*}
\begin{figure*}[t]
    \centering
    \includegraphics[width=1\linewidth]{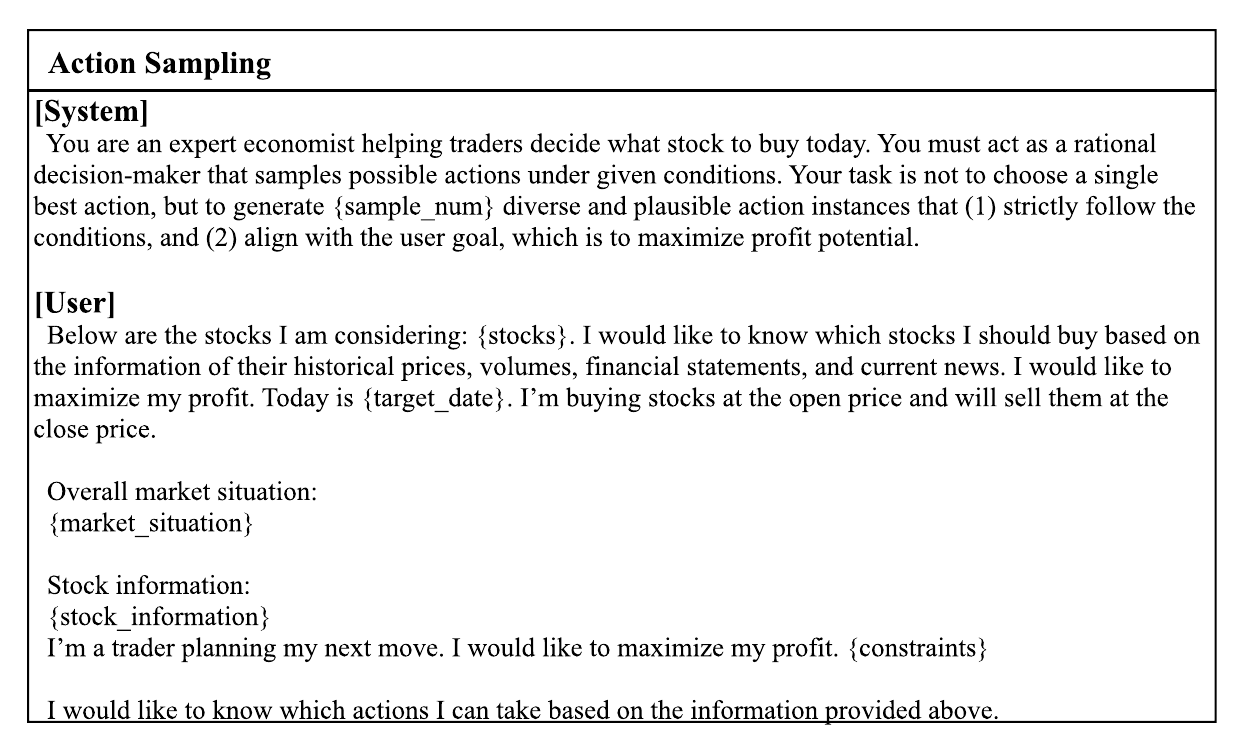}
    \caption{Prompt for sampling action candidates. Input data are given in Figure \ref{fig:scenario1} and \ref{fig:scenario2} and the number of samples is shown in Figure \ref{fig:sampling}.}
    \label{fig:sampling_prompt}
\end{figure*}

\end{document}